\documentclass[10pt,journal,compsoc]{IEEEtran}
\usepackage{times}
\usepackage{graphicx}
\usepackage{amssymb}
\usepackage{multirow}
\usepackage{bbm}
\usepackage{rotating}
\usepackage{array}
\usepackage{threeparttable}

\usepackage{subfigure}
\usepackage{amsmath}
\usepackage{bm}
\usepackage{url}
\usepackage{color}
\usepackage{algorithm}
\usepackage{algorithmic}
\usepackage[none]{hyphenat}

\ifCLASSOPTIONcompsoc
  \usepackage[nocompress]{cite}
\else
  \usepackage{cite}
\fi

\begin{document}
\title{Meta Balanced Network for Fair Face Recognition}

\author{Mei Wang, Yaobin Zhang, Weihong Deng
\IEEEcompsocitemizethanks{\IEEEcompsocthanksitem Mei Wang and Yaobin Zhang are with the Pattern Recognition and Intelligent System Lab., School of Artificial Intelligence, Beijing University of Posts and Telecommunications, Beijing, China. E-mail: \{wangmei1,zhangyaobin\}@bupt.edu.cn.
\IEEEcompsocthanksitem Weihong Deng is with the Pattern Recognition and Intelligent System Lab., School of Artificial Intelligence, Beijing University of Posts and Telecommunications, Beijing, China, and also with Key Lab. of Trustworthy Distributed Computing and Service, Ministry of Education, Beijing University of Posts and Telecommunications, Beijing, China. E-mail: whdeng@bupt.edu.cn.}

\thanks{(Corresponding author: Weihong Deng)}}

\markboth{Journal of \LaTeX\ Class Files,~Vol.~14, No.~8, August~2015}%
{Shell \MakeLowercase{\textit{et al.}}: Bare Demo of IEEEtran.cls for Computer Society Journals}

\IEEEtitleabstractindextext{%
\begin{abstract}
Although deep face recognition has achieved impressive progress in recent years, controversy has arisen regarding discrimination based on skin tone, questioning their deployment into real-world scenarios. In this paper, we aim to systematically and scientifically study this bias from both data and algorithm aspects. First, using the dermatologist approved Fitzpatrick Skin Type classification system and Individual Typology Angle, we contribute a benchmark called Identity Shades (IDS) database, which effectively quantifies the degree of the bias with respect to skin tone in existing face recognition algorithms and commercial APIs. Further, we provide two skin-tone aware training datasets, called BUPT-Globalface dataset and BUPT-Balancedface dataset, to remove bias in training data. Finally, to mitigate the algorithmic bias, we propose a novel meta-learning algorithm, called Meta Balanced Network (MBN), which learns adaptive margins in large margin loss such that the model optimized by this loss can perform fairly across people with different skin tones. To determine the margins, our method optimizes a meta skewness loss on a clean and unbiased meta set and utilizes backward-on-backward automatic differentiation to perform a second order gradient descent step on the current margins. Extensive experiments show that MBN successfully mitigates bias and learns more balanced performance for people with different skin tones in face recognition. The proposed datasets are available at http://www.whdeng.cn/RFW/index.html.
\end{abstract}

}

\maketitle

\IEEEraisesectionheading{\section{Introduction}\label{sec:introduction}}
\IEEEPARstart{R}{ecently}, with the emergence of deep convolutional neural networks (CNN) \cite{krizhevsky2012imagenet,simonyan2014very,szegedy2015going,he2016deep,hu2017squeeze}, research focus of face recognition (FR) has shifted to deep-learning-based approaches \cite{wang2018deep,sun2014deep,schroff2015facenet} and the accuracy was dramatically boosted to above 99.80\% on the Labeled Faces in the Wild (LFW) dataset \cite{huang2007labeled}. However, the recognition accuracy is not only aspect to attend when designing learning algorithms. As a growing number of applications based on FR have integrated into our lives, its potential for unfairness is raising alarm. For example, Amazon's Rekognition Tool incorrectly matched the photos of 28 U.S. congressmen with the faces of criminals, especially the error rate was up to 39\% for Black faces. According to these reports \cite{garvie2016perpetual,MITREVIEW}, FR system seems discriminative based on classes like race, demonstrating significantly different accuracy when applied to different groups. Such bias can result in mistreatment of certain demographic groups, by either exposing them to a higher risk of fraud, or by making access to services more difficult.  Consequently, there is an increased need to guarantee fairness for automatic systems and prevent discriminatory decisions.

\begin{table*}[htbp]
    \setlength{\abovecaptionskip}{0cm}
	\setlength{\belowcaptionskip}{-0.2cm}
    \caption{Bias with respect to skin tone in commercial APIs and SOTA FR algorithms. Verification accuracies (\%) on our IDS-8 are given. Skin gradually darkens with the increase of tone value (from I to VIII). We make the best results bold, and make the worst in red.}    \label{racial bias of 8bin}
	\begin{center}
	\footnotesize
    \begin{threeparttable}
    \setlength{\tabcolsep}{4mm}{
	\begin{tabular}{c|cccccccc}
		\hline
        \multirow{2}{*}{Model} & \multicolumn{8}{c}{The skin tone of IDS-8} \\
                               & I & II & III & IV & V & VI & VII & VIII  \\ \hline \hline
         Microsoft \cite{azure} &\textcolor{red}{88.67} & \textcolor{red}{87.15} & \textcolor{red}{80.20} & \textcolor{red}{78.45} & \textcolor{red}{83.13} & \textcolor{red}{82.09} & \textcolor{red}{76.10} & \textcolor{red}{74.90}   \\
         Face++ \cite{Face++} &  \textbf{93.68} &  \textbf{93.89} &  \textbf{92.45} &  \textbf{92.49} &  \textbf{88.58} &  \textbf{89.36} &  \textbf{88.23} &  \textbf{87.40} \\
         Baidu \cite{Baidu} &90.52 & 88.04 & 90.52 & 89.68 & 86.72 & 87.19 & 78.18 & 78.62  \\
         Amazon \cite{amazon} &91.22 & 90.12 & 84.96 & 85.13 & 86.85 & 88.40 & 85.79 & 86.36  \\ \hline
         mean & 91.02 & 89.80 & 87.03 & 86.44 & 86.32 & 86.76 & 82.07 & 81.82  \\ \hline
         Center-loss \cite{wen2016discriminative}  & \textcolor{red}{87.59} & \textcolor{red}{87.14} & \textcolor{red}{79.21} & \textcolor{red}{77.87} & \textcolor{red}{81.90} & \textcolor{red}{83.68} &	 \textcolor{red}{78.55} & \textcolor{red}{77.51} \\
         Sphereface \cite{liu2017sphereface} & 90.59 & 90.71 & 82.75 & 82.42 & 85.50 & 88.06 & 81.95 & 81.82  \\
         Arcface\tnote{1} \cite{deng2018arcface}  &  \textbf{92.33} &  \textbf{91.70} & 83.97 & 83.05 &  \textbf{87.46} &  \textbf{88.40} &  \textbf{84.07} &  \textbf{83.23}  \\
         VGGface2\tnote{1} \cite{cao2017vggface2} & 90.22 & 89.66 &  \textbf{84.93} &  \textbf{83.82} & 86.03 & 87.29 & 83.37 & 82.72  \\ \hline
         mean & 90.18 & 89.80 & 82.71 & 81.79 & 85.22 & 86.86 & 81.98 & 81.32  \\ \hline
	\end{tabular}}
    \begin{tablenotes}
     \item[1] Arcface is a ResNet-34 model trained with CASIA-Webface. VGGFace2 is a SeNet model trained with VGGFace2.
    \end{tablenotes}
    \end{threeparttable}
    \end{center}
\end{table*}

\begin{figure*}[htbp]
\centering
\includegraphics[width=18cm]{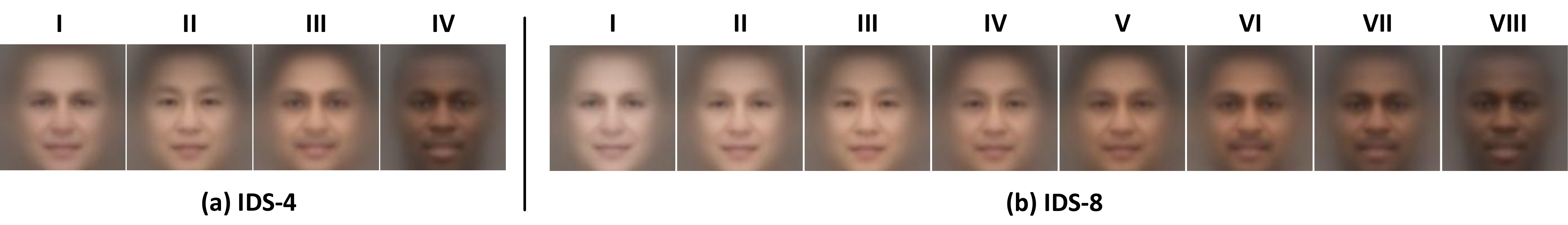}
\caption{ The average faces of different skin tone bins of IDS database which consist of the average pixel values computed from aligned faces.} 
\label{fig1}
\end{figure*}

Although several studies \cite{phillips2011other,klare2012face} have uncovered such discrimination in non-deep FR algorithms, there are still no sufficient research efforts in deep learning era. Without a dataset that has demographic labels for various people, one cannot systematically examine the inappropriate biases in trained models. To facilitate the research towards this issue, in this paper, we have done the first step to overcome the major obstacle. Considering race labels are unstable, we decided to use skin tone as a more precise and scientific label. With the help of the Fitzpatrick Skin Type classification system \cite{fitzpatrick1988validity} and Individual Typology Angle \cite{merler2019diversity,chardon1991skin}, a new test dataset, called Identity Shades (IDS), is constructed which is phenotypically balanced on the basis of skin tone, as shown in Fig. \ref{fig1}. It can be used to fairly evaluate FR algorithms across faces with different skin tones. Based on experiments on IDS, we find that both commercial APIs and state-of-the-art (SOTA) algorithms indeed suffer from bias: the error rates on dark-skinned faces are about two times of the light-skinned ones, as shown in Table 1. Moreover, we demonstrated that this bias comes from both data and algorithm aspects. Hence, further research efforts on data and algorithms are requested to eliminate this bias.

A major driver of bias in FR is the training data. Large-scale datasets, such as CASIA-WebFace \cite{yi2014learning}, VGGFace2 \cite{cao2017vggface2} and MS-Celeb-1M \cite{guo2016ms}, are typically constructed by scraping websites like Google Images. Such data collecting methods can unintentionally produce data that encode biases with respect to skin tone. Thus, social awareness must be brought to the building of datasets for training. In this work, we take steps to ensure such datasets are diverse and do not under represent particular skin tone groups by constructing two new training datasets, i.e., BUPT-Globalface and BUPT-Balancedface dataset. One is built up according to the approximated distribution of average skin tones around the world, and the other strictly balances the number of samples in skin tone.

Another source of bias can be traced to the algorithms. The state-of-the-art (SOTA) face recognition methods, such as Cosface \cite{wang2018cosface} and Arcface \cite{deng2018arcface}, apply a fixed margin between classes to maximize overall prediction accuracy for the training data. 
However, maximizing overall accuracy might come at the expense of the under-represented populations and leads to poor performance for those subjects, which even amplifies the biases in data. To address this problem, the algorithms must trade-off the specific requirements of margins of various groups of people, and set adaptive margins for faces with different skin tones to produce more equitable recognition performance. But this demographic bias is a complex problem caused by many latent factors, including but not limited to the quantity, it's quite difficult to manually preset or tune these margins by heuristic strategies like cross-validation, naturally conducting efficiency issue and difficulty in practical implementations.

In our paper, we propose a novel meta-learning algorithm, named Meta Balanced Network (MBN), which automatically learns adaptive margin for each skin tone group based on their gradients from meta data. We additionally use a small but unbiased validation set, i.e., meta data, to guide training in our MBN. The meta data can help the model to distinguish fairness from bias and learn high-level causal relationships in bias. During training, we treat model optimization as the objective of inner algorithm trained by training data, and treat margin optimization as the objective of outer algorithm trained by meta data. The outer algorithm can evaluate the bias of the learned model on meta data and dynamically output margins in adaptive margin loss function; and the inner algorithm aims to optimize the model guided by adaptive margin loss function on training data such that the model performs well and fairly across faces with different skin tones. A specific backward-on-backward differentiation enables us to connect disjoint process between inner algorithm and outer algorithm and transmit meta gradient from meta loss to training loss to update the margins. This bilevel-optimization realizes a mutual amelioration between automatically tuning margin parameters involved in adaptive margin loss and learning suitable model parameters leading to balanced performance.

Our contributions can be summarized into three aspects.

1) We construct and release three large-scale in-the-wild datasets, i.e., IDS, BUPT-Globalface and BUPT-Balancedface, which are balanced by skin tone. They are the first series of databases for studying bias with respect to skin tone in both training and evaluating aspects. Based on the extensive experiments on them, we not only measure the bias in commercial systems and deep recognition models, but also validate the bias comes from both data and algorithm.

2) To mitigate the algorithmic bias, we propose to trade off the specific requirements of margins of various people. Our novel MBN method leverages an additional small meta set to automatically learn the optimal margins by utilizing backward-on-backward automatic differentiation to take a second order gradient pass in meta-optimization. To the best of our knowledge, this is the first time that meta learning is used to solve the fairness of recognition problem.

3) Extensive experiments on the Globalface, Balancedface, and IDS datasets show that our meta balanced network (MBN) shows more balanced performance on different skin tone subjects than traditional sample re-weighting method, adversarial attribute removal method and our recently proposed reinforcement marginal learning method \cite{wang2019mitigate}.

The remainder of this paper is structured as follows. In the next section, we discuss the studies of bias, and also review the related approaches of debiasing algorithms and meta learning. Then, we introduce the details of our databases in Section III. In Section IV, we propose the meta balanced network, i.e., MBN, to learn balanced performance for different skin tone groups. In Section V, experimental results on IDS are shown and validate that the existence and cause of such bias. Then we evaluate the effectiveness of our MBN method. Finally, we conclude and discuss future work.

\section{Related work}

\subsection{Bias with respect to skin tone in face recognition}

Some studies \cite{pmlr-v81-buolamwini18a,raji2019actionable,aminiuncovering} have raised concerns about face analysis systems, e.g., gender classification, being biased based on some classes like race. The study of such bias in face recognition, likewise, has a nearly 30-year history that converges on the following aspects.

\textbf{Algorithms.} There are published studies \cite{grother2010report,phillips2003face,phillips2011other,klare2012face} analyzing the performances of face recognition algorithms over demographic groups and uncovering that these algorithms suffer from bias. The 2002 NIST Face Recognition Vendor Test (FRVT) \cite{phillips2003face} showed that non-deep FR algorithms have different recognition accuracies depending on demographic cohort, such as skin color, age and gender. Phillips et al. \cite{phillips2011other} suggested that training and testing on different races results in severe performance drop. Klare et al. \cite{klare2012face} evaluated six different FR algorithms on three skin tone groups, and concluded that the Black cohorts are more difficult to recognize for all matchers. The FRVT 2019 \cite{grother2019face} showed the demographic bias of over 100 face recognition algorithms. Krishnapriya et al. \cite{vangara2019characterizing,krishnapriya2020issues} found that darker-skinned subjects have a higher false match rate, and lighter-skinned ones have a higher false nonmatch rate through experiments on MORPH dataset \cite{ricanek2006morph}.

\textbf{Datasets.} In non-deep learning era, some training and testing datasets are constructed for studying and measuring this demographic bias in FR algorithms. Klare et al. \cite{klare2012face} collected mug shot face images of Whites, Blacks and Hispanics from the Pinellas County Sheriff's Office (PCSO). Furl et al. \cite{furl2002face} collected images of Caucasians, Asians, Africans, Indians and  Hispanics from the FERET database \cite{Phillips1998The}. Phillips et al. \cite{phillips2011other} utilized the images of FRVT 2006 \cite{beveridge2008focus} with a sufficient number of Caucasian and East Asian faces to conduct cross training and matching on White and Asian faces. However, few studies focus on bias in deep era because benchmark datasets are lacking to systematically examine the inappropriate biases in trained models. In this paper, we take a step in this direction by releasing three datasets. Considering race labels are unstable as we explained in Section \ref{data}, we decided to use skin tone as a more precise label.

\subsection{Debiasing algorithms}

There are some works that seek to introduce fairness into machine learning pipelines and mitigate data bias. 

\textbf{Unbalanced-training method.} Sample reweighting methods \cite{sattigeri2018fairness,calmon2017optimized} have been believed to be an effective way in addressing class imbalance problem, which decreases weights for the over-sized classes and imposes larger weights on rarer data. For example, DB-VAE \cite{aminiuncovering} used the learned latent distributions to re-weight the importance of data points. Other methods \cite{wang2019mitigate,gong2021mitigating} mitigate the bias via model regularization, taking into consideration of the fairness goal in the overall model objective function. For example, RL-RBN \cite{wang2019mitigate} formulated the process of finding the optimal margins for non-White people as a Markov decision process and employed deep Q-learning to learn policies based on large margin loss.

\textbf{Attribute removal method.} By confounding or removing demographic information of faces, attribute removal methods \cite{alvi2018turning,mirjalili2018gender,othman2014privacy} make a great contribution to mitigating the hidden, and potentially unknown, biases within training data. Alvi et al. \cite{alvi2018turning} applied a confusion loss to make a classifier fail to distinguish attributes of examples so that multiple spurious variations are removed from the feature representation. SensitiveNets \cite{morales2019sensitivenets} minimized sensitive information in triplet loss while retaining recognition ability. Gong et al. \cite{gong2020jointly} debiased face recognition by disentangling features related to demographics and identity.

\textbf{Domain adaptation.} Some papers \cite{wang2019racial,kan2015bi,kan2014domain,Zhang_2017_CVPR} proposed to investigate bias problem from a domain adaptation point of view and attempted to learn domain-invariant feature representations to mitigate bias across domains. IMAN \cite{wang2019racial} simultaneously aligned global distribution to decrease gap at domain-level, and learned the discriminative target representations at cluster level. Kan \cite{kan2015bi} directly converted the White faces to non-White domain in the image space with the help of sparse reconstruction coefficients.

\subsection{Meta learning}

The field of meta learning, or learning to learn, has seen a dramatic rise in interest in recent years. Recent meta-learning studies concentrate on: 1) learning a good weight initialization for fast adaptation on a new task \cite{finn2017model,nichol2018first,sun2019meta,rajeswaran2019meta}. For example, Ravi and Larochelle \cite{ravi2016optimization} proposed to learn the few-shot optimization algorithm with an LSTM-based meta-learner. MAML \cite{finn2017model} and its variants Reptile \cite{nichol2018first} simplified the above meta-learner model and only learned the initial learner parameters. MFR \cite{guo2020learning}  synthesized the source/target domain shift with a meta-optimization objective to improve generalization in face recognition. 2) learning an adaptive weighting scheme \cite{ren2018learning,shu2019meta}. For example, to address label noise and long-tail problem, L2RW \cite{ren2018learning} adopted a novel meta-learning algorithm that learns to assign weights to training data based on their gradient directions. Different from learning weights explicitly in L2RW \cite{ren2018learning}, Shu et al. \cite{shu2019meta} proposed to utilize meta learning to learn a weight function, i.e., Meta-Weight-Net, which can learn the weights in a more stable way. Our approach is most related to L2RW \cite{ren2018learning}. However, 1) L2RW used meta learning to re-weight training samples, while our MBN learns adaptive margin. 2) Fairness is not taken into consideration in L2RW which only aims to improve total performance on long-tail data. Thus, L2RW utilized traditional classification loss as meta loss, while our MBN proposes a novel meta skewness loss to evaluate the bias of model during training such that model is required to learn fair representations on meta data. 

\section{Databases for unbiased training and fairness evaluation}

In deep face recognition, few studies focus on bias or fairness problem with respect to skin tone because so few balanced training and testing datasets are available. The deep networks are often designed to fit biased training data, and thus would naturally replicate the biases already existent in data; and test sets contained the same biases as the training set might fail to unveil the unfairness problem of trained models. To address this issue, we contribute two training and one testing databases, i.e., BUPT-Globalface, BUPT-Balancedface and IDS, to enable researchers to go deep into this issue.

\subsection{Rationale for skin tone labeling}  \label{data}

Although race or ethnicity based categories have been used in some fields \cite{friedler2016possibility,machine2016julia}, they seem unstable in computer vision. First, the drawing boundaries between distinct races are complex and even confused. Thus, inconsistent definition and use have been chief problems with the race concept, particularly as the prevalence of admixture increases across populations. Second, the concept of race is a social construct without biological and physiological basis \cite{yudell2016taking,pmlr-v81-buolamwini18a}. Subjects' phenotypic features, e.g., skin color, eye and nose shapes, can vary widely within a racial or ethnic category. As such, explicitly annotating and building models on top of race or ethnicity risks to perpetuate bias and potentially dangerous outcomes. To scientifically study this demographic bias, we follow the works of Buolamwini et al. \cite{pmlr-v81-buolamwini18a} and Merle et al. \cite{merler2019diversity}, and use skin tone as a more visually precise label.

\subsection{Data collection and annotation}

We download face images of diverse regions around world from Google according to one-million FreeBase celebrity list \cite{freebase}, and clean them both automatically and manually in the similar way as other FR training datasets, e.g., VGGface2 \cite{cao2017vggface2} and Megaface \cite{kemelmacher2016megaface}. Then, the skin tones of these downloaded images are estimated. We utilized Individual Typology Angle (ITA) to automatically measure skin tones of images following the work of Merle et al. \cite{merler2019diversity}. Note that the skin tone annotations of testing images are further manually checked using Fitzpatrick Skin Type classification system \cite{fitzpatrick1988validity,pmlr-v81-buolamwini18a}. Thus, images can be divided into several bins based on a mapping to skin tone. We group people into 8 bins, named ``Tone I-VIII". Skin gradually darkens with the increase of tone value, that is, people with Tone-I have the lightest skin color and those with Tone-VIII have the darkest skin tone. After that, we construct our training and testing datasets, i.e., Globalface-8, Balancedface-8 and IDS-8, using these annotated images. Our recently published RFW testing dataset \cite{wang2019racial} is constructed by the similar way and is divided into 4 bins. For consistency, we call it IDS-4 throughout the paper.

\subsection{BUPT-Globalface and BUPT-Balancedface}

\begin{figure}[htbp]
\centering
\subfigure[4 skin bins]{
\label{ratio_4bin} 
\includegraphics[width=3.8cm]{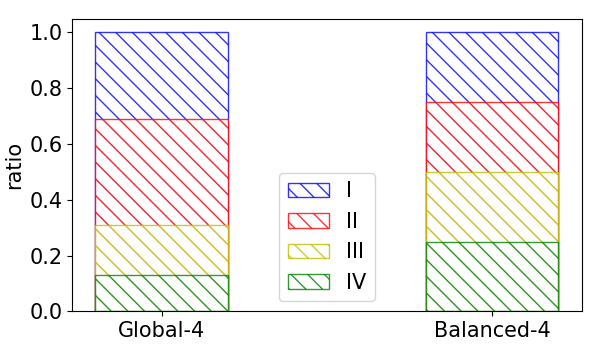}}
\hspace{0cm}
\subfigure[8 skin bins]{
\label{ratio_8bin} 
\includegraphics[width=3.8cm]{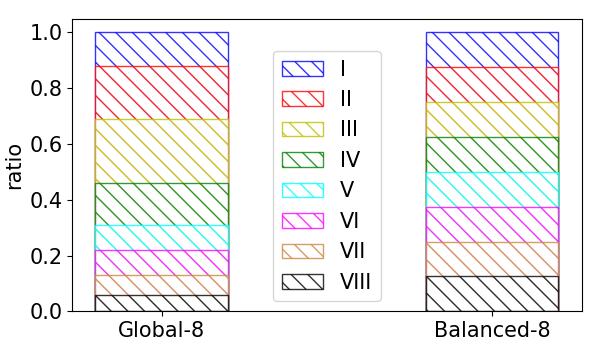}}
\caption{ The skin tone distributions of BUPT-Globalface and BUPT-Balancedface datasets. }
\label{ratio1} 
\end{figure}

To remove bias from data aspect and represent people of different skin tones equally, we construct two training datasets, i.e., BUPT-Globalface and BUPT-Balancedface. Globalface contains 2M images from 38K celebrities in total and its distribution is approximately the same as real distribution of world's population. Balancedface dataset contains 1.3M images from 28K celebrities and is approximately balanced with respect to skin tone. Specifically, Globalface-4 and Balancedface-4 are divided into 4 skin bins, named ``Tone I-IV", based on a mapping to skin tone; while the images in Globalface-8 and Balancedface-8 are divided into 8 bins, i.e., ``Tone I-VIII", as shown in Fig. \ref{ratio1}.

\subsection{Identity shades dataset: IDS}

\begin{figure*}[htbp]
\centering
\subfigure[yaw pose]{
\label{fig10a} 
\includegraphics[width=3.8cm]{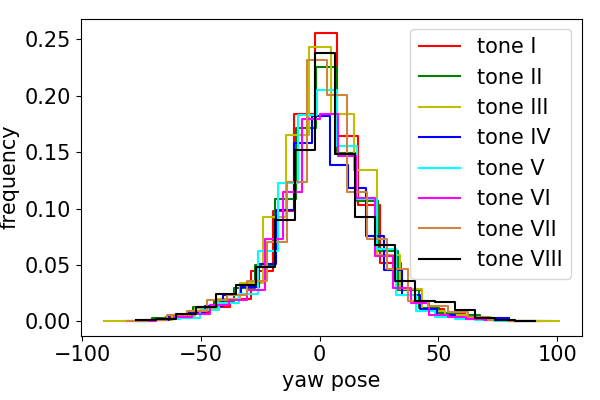}}
\hspace{0cm}
\subfigure[pitch pose]{
\label{fig10b} 
\includegraphics[width=3.8cm]{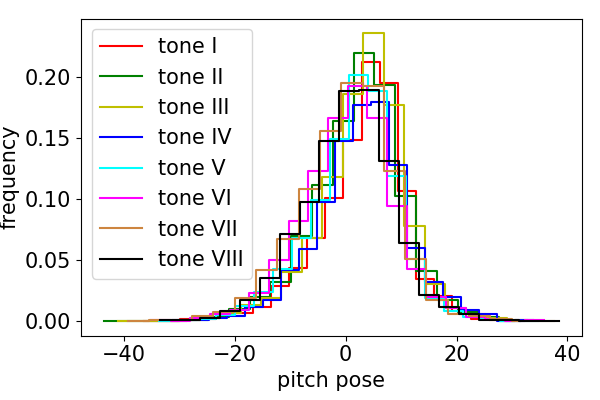}}
\hspace{0cm}
\subfigure[age]{
\label{fig10c} 
\includegraphics[width=3.8cm]{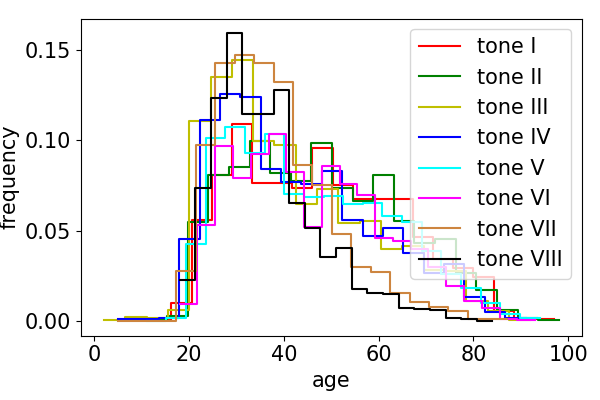}}
\hspace{0cm}
\subfigure[gender]{
\label{fig10d} 
\includegraphics[width=3.8cm]{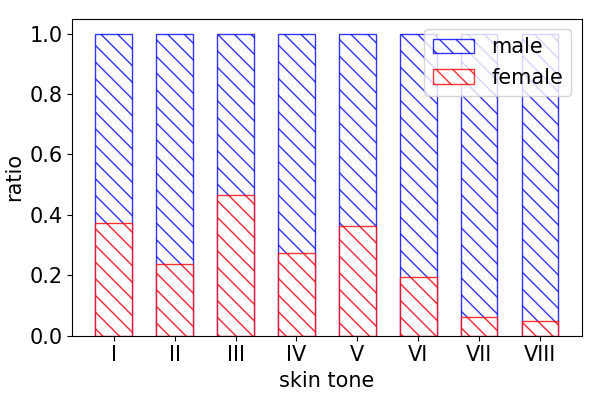}}
\caption{ IDS statistics. We show yaw pose, pitch pose, age and gender distribution of eight testing bins of IDS-8.}
\label{fig10} 
\end{figure*}

IDS-4 database is a testing set which contains four skin bins. Each bin contains about 10K images of 3K individuals for face verification. For easy comparison, 6K difficult pairs of images are selected for each bin. Similarly, IDS-8 consists of 8 skin bins and contains 3K difficult pairs of images per bin. All of these images have been carefully and manually cleaned. For the performance evaluation, we recommend to use both the biometric receiver operating characteristic (ROC) curve and LFW-like protocol. Specifically, ROC curve, which aims to report a comprehensive performance, evaluates algorithms on all pairs of identities (about 14K positive vs. 50M negative pairs for IDS-4). In contrast, LFW-like protocol facilitates easy and fast comparison between algorithms with selected 6K (or 3K) pairs of images for IDS-4 (or IDS-8). Further, inspired by the ugly subset of GBU database \cite{Phillips2012The}, we have selected the ``difficult" pairs (in term of cosine similarity) to avoid the saturated performance to be easily reported.

\begin{figure}[htbp]
\centering
\includegraphics[width=8cm]{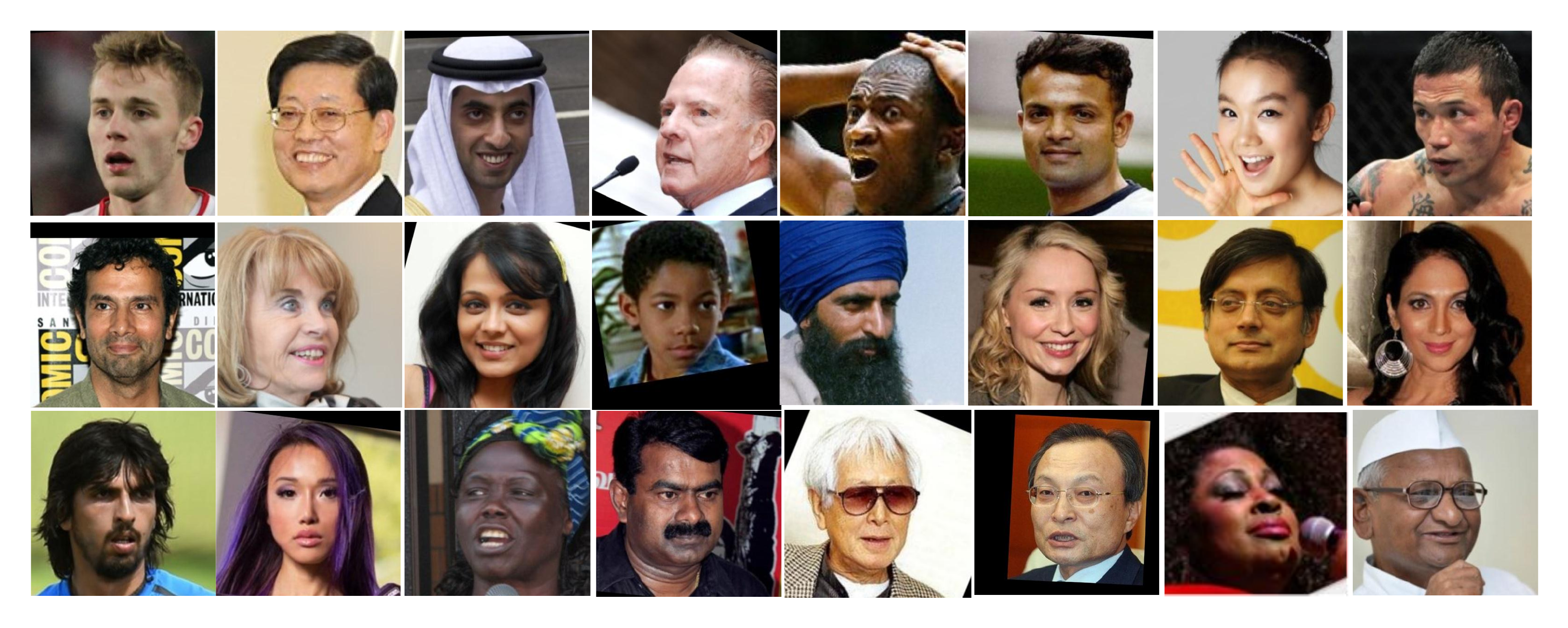}
\caption{ Some images examples of our IDS. One can see from the figure that faces in our dataset are from diversity regions around world and exhibit variability in factors such as pose, age and expression. }
\label{example_data}
\end{figure}

Some image examples of our IDS are shown in Fig. \ref{example_data}. In IDS, the images of each skin bin are randomly collected without any preference, there is no significant difference between different bins besides skin color, and thus they are suitable to fairly measure the bias with respect to skin tone. We have validated that, across varying bins, their distributions of pose, age, and gender are similar. As evidence, the detailed distributions measured by Face++ API are shown in Fig. \ref{fig10}. 

\section{Meta balanced network}

In our meta balanced network, we introduce the idea of adaptive margin into fairness problem. The optimal margins are automatically learned by meta learning to trade-off the specific requirements of people with different skin tones. Compared with traditional sample re-weighting method, our MBN can balance the feature scatter of different skin tone groups in feature space instead of just imposing different weights on loss. In face recognition which is a fine-grained and open-set classification problem, MBN can obtain better generalization ability and more balanced performance across various skin tone subjects.

\subsection{Adaptive margin loss}

\begin{figure*}[htbp]
\centering
\includegraphics[width=16cm]{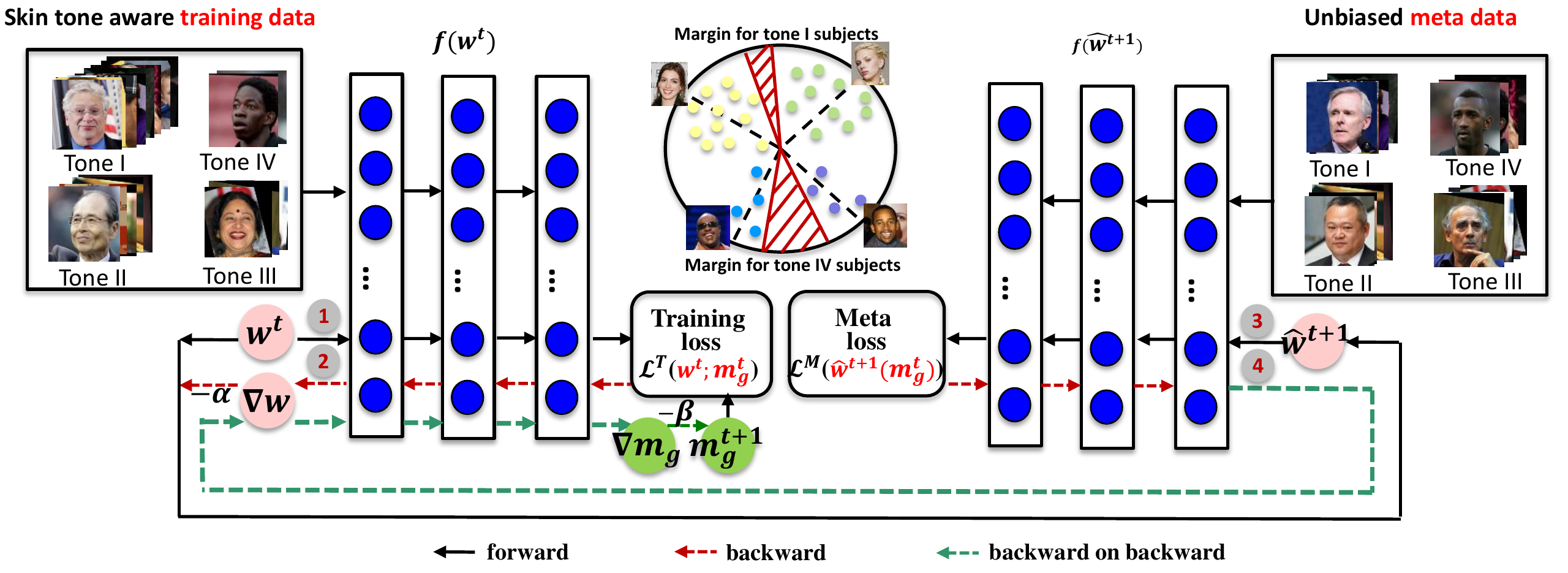}
\caption{ An illustration of our method. At iteration step $t$, \textbf{Margin parameters learning:} first, given a mini-batch of training samples, we use SGD to update the model parameters $\widehat{w}^{t+1}$ on the basis of the current parameters $w^{t}$ and margins $m_g^{t}$ (step 1 and 2). Then, a mini-batch of meta samples is sent to the updated model (step 3). Guided by our meta skewness loss, a meta-gradient (high-order gradient) is back-propagated from meta data to training data, respectively, to update the margins $m_g^{t+1}$ (step 4). \textbf{Model parameters learning:} with a fixed margin for Tone-I subjects and the updated margins $m_g^{t+1}$ for other faces, we use adaptive margin loss and training data to update the model parameters $w^{t+1}$ such that it performs fairly across people with different skin tones. }
\label{MBN}
\end{figure*}

Although large margin losses, e.g., Cosface \cite{wang2018cosface} and Arcface \cite{deng2018arcface}, successfully improve feature discrimination, and get better performance on FR benchmarks, they still fail to obtain balanced representations on different skin tone bins, as shown in Table \ref{racial bias of 8bin}. The decision boundary in these losses is assigned the same margin without considering the requirements of people leading to biased performance. In order to address this bias problem and prevent unintended discrimination in existing face recognition algorithms, we suggest that algorithms must trade-off the specific requirements of margins of various groups of people to control their fairness extents under different skin-tone distributions. Therefore, we introduce the idea of adaptive margin into fairness problem with respect to skin tone, and the details are as follows. Note that we take 4 skin bins as examples to describe our method and 8 skin bins can be processed by the similar way.

Since lighter-skinned (Tone-I) subjects are overwhelmingly dominant in numbers and perform best in existing FR datasets, we make lighter-skinned group as the benchmark (anchor) in our paper by remaining the margins of Tone-I subjects unchanged. Optimal margins are learned adaptively for other darker-skinned groups (II-IV) in order to minimize the skewness between lighter- and darker-skinned subjects. In our MBN, we replace the fixed margin in Arcface \cite{deng2018arcface} by a skin tone related and training step related parameter $m_{g}(t)$, where $m_g \in \{m_{II},m_{III},m_{IV}\}$ is the margin corresponding to darker-skinned group $g$ and $t$ represents the stage of the training. The proposed adaptive margin loss function can be formulated as follows:

\begin{footnotesize}
\begin{equation}
\begin{split}
&L_j^{T(arc)}=-log\frac{e^{s\left ( cos\left ( \theta_{y_{j}} +\lambda _{g_j}(t) \right ) \right )}}{e^{s\left ( cos\left ( \theta_{y_{j}} +\lambda _{g_j}(t) \right ) \right )}+\sum_{i=1,i\neq y_{j}}^{c}e^{s\cdot cos\theta _{i}}} \label{RBN-arc}\\
&where,\ \lambda _{g_j}(t) =\left\{
             \begin{array}{lll}
             m, & if \ g_j= Tone \ I & \\
             m_{g}(t), & otherwise & \\
             \end{array}
\right.
\end{split}
\end{equation}
\end{footnotesize}
where $g_j$ is the skin-tone label of $j$-th sample. $\theta _{i}$ is the angle between the weight $W_{i}$ and the feature $z_{j}$. $z_j\in \mathbb{R}^{d}$ denotes the deep feature of the $j$-th sample, belonging to the $y_{j}$-th class, and $W_i\in \mathbb{R}^{d}$ denotes the $i$-th column of the weight $W\in \mathbb{R}^{d\times n}$. $c$ is the number of classes and $s$ is the scale factor. The similar modification can be made for Cosface \cite{wang2018cosface} as follows:

\begin{footnotesize}
\begin{equation}
\begin{split}
&L_j^{T(cos)}=-log\frac{e^{s\left ( cos\left ( \theta_{y_{j}}  \right ) -\lambda _{g_j}(t) \right )}}{e^{s\left ( cos\left ( \theta_{y_{j}}  \right ) -\lambda _{g_j}(t) \right )}+\sum_{i=1,i\neq y_{j}}^{c}e^{s\cdot cos\theta _{i}}} \label{RBN-cos}\\
&where,\ \lambda _{g_j}(t) =\left\{
             \begin{array}{lll}
             m, & if \ g_j= Tone \ I  & \\
             m_{g}(t), & otherwise& \\
             \end{array}
\right.
\end{split}
\end{equation}
\end{footnotesize}

So the key problem is to set optimal margin $m_g(t)$ for each darker-skinned bin $g$ to minimize the skewness between lighter-skinned and darker-skinned subjects. However, bias is a complex problem caused by many latent factors, including but not limited to the quantity. Instead of manually presetting or tuning them by cross-validation, we provide the following algorithm to adaptively learn these margins, by borrowing the idea of recent meta-learning techniques \cite{finn2017model,ren2018learning,shu2019meta}.

\subsection{Meta margin learning} \label{dqn}

Our meta margin learning consists of two parts: 1) margin optimization which is responsible for setting training loss (adaptive margin loss) for model optimization by outputting appropriate margin $m_g$; 2) model optimization which optimizes the network by training loss. We aim to learn the optimal training loss such that the network guided by this can perform fairly across different skin tone subjects. To accomplish this, first, we design a skewness loss to evaluate the fairness of learned model by which the margin parameters can be optimized. However, when the training set is biased, the skewness loss imposed on training data would have the wrong perception of bias. According to \cite{ren2018learning}, without a proper definition of an unbiased set, solving the training set bias problem is inherently ill-defined. Therefore, we propose to utilize a clean and balanced meta set to evaluate bias of the learned model and use a large but biased training set to optimize network. Third, to connect disjoint process between training and meta-evaluating, we utilize backward-on-backward automatic differentiation in meta-optimization such that we can transmit meta gradient from meta skewness loss to training loss to update the margins. Benefiting from this bilevel-optimization, we realize a mutual amelioration between model and margin optimization. Then, we introduce the details of our meta learning algorithm.

\textbf{Model optimization.} Let $\mathcal{X}\in \mathbb{R}^{d}$ be the image space, $\mathcal{Y} = \{1, 2,..., c\}$ be the identity label space, and $\mathcal{G} = \{1, 2,..., k\}$ be the skin-tone label space. $D_{train}=\{x_i, y_i, g_i\} _{i=1}^N$ denotes a large biased training set, where $N$ is the number of training samples. $D_{train}$ is divided into 4 bins based on a mapping to skin tone, i.e., $D_{train}^{g}=\{x_i^{g}, y_i^{g}\} _{i=1}^{N_g}$, where $g\in\{I,II,III,IV\}$. 
Let $f(x; w)$ be our neural network model, and $w$ be the model parameters. The training of model $f(x; w)$ is an optimization process that discovers a good model parameter $w^{*}$ by minimizing our adaptive margin loss on the training data. Our adaptive margin loss can be formulated as $\frac{1}{N}\sum_{j=1}^{N}L_j^{T(arc)}(w;m_{g_{j}})$ or $\frac{1}{N}\sum_{j=1}^{N}L_j^{T(cos)}(w;m_{g_{j}})$. For notation convenience, we denote that $L_j^T(w;m_{g_{j}})=L_j^T(f(x_j;w);y_j;m_{g_{j}})$. So the model optimization can be formulated as follows:
\begin{equation}
w^{*}(m_g)=arg\mathop{min}\limits_{w}\frac{1}{N}\sum_{j=1}^{N}L_j^{T}(w;m_{g_{j}}) \label{model}
\end{equation}
where $m_g=\{m_{II},m_{III},m_{IV}\}$ denotes the margins of darker-skinned groups and $g_{j}$ is the skin tone label of $j$-th sample. Note that $m_g$ can be understood as training hyper-parameters and learned automatically during training.

\textbf{Margin optimization.} To cater for different requirements of different people and different status of model training, we should learn different margins for different skin tone bins at each training step $t$ to balance the performance.
Assume that we additionally have a small meta-data set $D_{meta}=\{x_i^v, y_i^v, g_i^v\} _{i=1}^M$ with clean labels and balanced data distribution, where $M$ is the number of meta-samples, and $M\ll N$.  $D_{meta}$ is also divided into 4 bins, i.e., $D_{meta}^{g}=\{x_i^{v,g}, y_i^{v,g}\} _{i=1}^{M/4}$, where $g\in\{I,II,III,IV\}$. The optimal selection of $m_g$ is based on the performance on meta set. 

We aim to learn the optimal margins $m_g^{*}$  based on this basic assumption: \emph{when testing the model performance on meta set, the model parameters $w^{*}(m_g^{*})$ trained by adaptive margin loss $L^T(w;m_g^{*})$ should reduce the bias (skewness) between lighter-skinned (I) and darker-skinned (II-IV) subjects in meta set}. Thus, we can then formulate a meta skewness loss minimization problem with respect to $m_g$ as:
\begin{equation}
m_g^{*}=arg\mathop{min}\limits_{m_g}L^{M}(w^{*}(m_g)) \label{update margin}
\end{equation}
where $m_g=\{m_{II},m_{III},m_{IV}\}$, and $L^{M}$ represents the meta skewness loss imposed on meta data and can reflect the skewness between lighter- and darker-skinned subjects in meta set. It consists of three parts: $B^{II}$, $B^{III}$ and $B^{IV}$:
\begin{equation}
L^M=B^{II}+B^{III}+B^{IV} \label{meta loss1}
\end{equation}
$B^{II}$ represents the bias (skewness) of feature scatter between Tone-I and Tone-II subjects, which is formulated as follows. $B^{III}$ and $B^{IV}$ can be computed by the same way as $B^{II}$.
\begin{equation}
B^{II}=\left | \frac{1}{M/4}\sum_{i=1}^{M/4}l^{II}_i-\frac{1}{M/4}\sum_{k=1}^{M/4}l^{I}_k \right | \label{meta loss2}
\end{equation}
where $l^{II}_i$ can reflect intra-class compactness and inter-class discrepancy of meta samples with skin tone II, and we use it to measure the model generalization on Tone-II subjects. $l^{I}_k $ can be computed by the same way as $l^{II}_i$ to measure the model generalization on subjects with skin tone I.

\begin{footnotesize}
\begin{equation}
l^{II}_i=exp\left ( \left \| f(x_{i}^{v,II})-f(x_{i,p}^{v,II}) \right \| _2^{2}-\gamma  \left \| f(x_{i}^{v,II})-f(x_{i,n}^{v,II}) \right \| _2^{2}\right ) \label{diff_indian}
\end{equation}
\end{footnotesize}
where $f(x_{i}^{v,II})$ represents the embedding of meta sample with skin tone II. For notation convenience, we denote that $f(x_{i}^{v,II})=f(x_{i}^{v,II};w)$. 
The subscripts $_{i,p}$ and $_{i,n}$ denote meta sample $x_{i}^{v,II}$'s hardest positive and negative samples, $\left \| f(x_{i}^{v,II})-f(x_{i,p}^{v,II}) \right \| _2^{2}$ is the L2-norm distance between sample $x_{i}^{v,II}$ and its positive sample $x_{i,p}^{v,II}$ to measure their similarity. $\gamma $ is the parameter for the trade-off between positive-pair distance and negative-pair distance. Therefore, $B^{II}$ represents the skewness of model performance between Tone-I and Tone-II subjects. Through minimizing $L^{M}$, we can learn the optimal margin $m_g^{*}$ in large margin loss $L^T(w;m_g^{*})$ which makes the learned model mitigate bias between lighter-skinned (I) and darker-skinned (II-IV) groups on meta data.

\begin{figure*}[htbp]
\centering
\includegraphics[width=14cm]{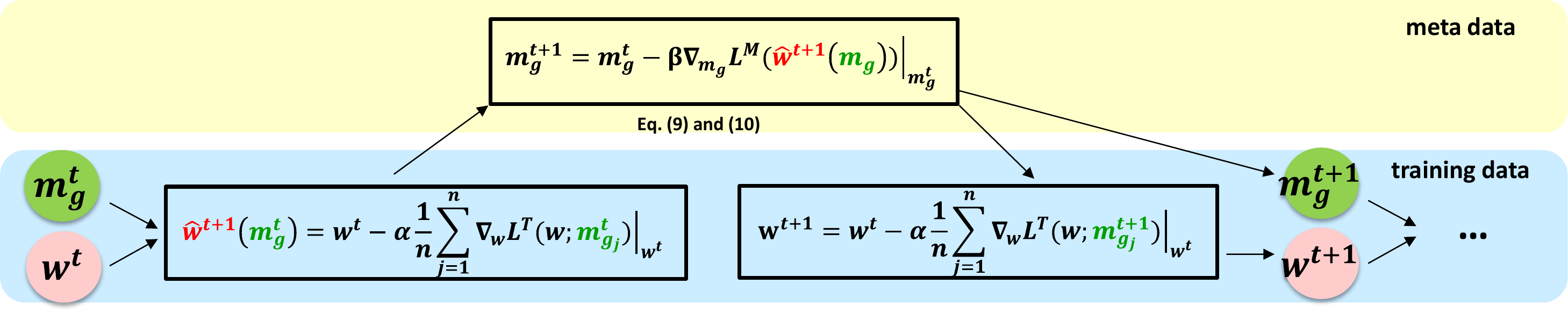}
\caption{Main flowchart of the proposed algorithm. Notice that $\widehat{w}^{t+1}$ here is a variable instead of a quantity, which makes $ \widehat{w}^{t+1}(m_g^t)$ a function of $m_g^t$ and the gradient in Eq. (\ref{margin update}) and (\ref{meta gradient}) be able to be computed by backward-on-backward differentiation.}
\label{iterative}
\end{figure*}

\subsection{Iterative meta-learning training strategy}

Calculating the optimal $w^{*}$ and $m_g^{*}$ requires two nested loops of optimization, which is expensive to obtain the exact solution \cite{franceschi2018bilevel}. Here we adopt an online approximation strategy \cite{finn2017model,ren2018learning} to jointly update both $w^{*}$ and $m_g^{*}$ in an iterative manner to guarantee the efficiency of the algorithm.

For most training of deep networks, SGD or its variants are used to optimize such loss functions. At every step $t$ of training, a mini-batch of training samples $\{(x_j,y_j ),1\leq j\leq n\}$ is sampled, where $n$ is the mini-batch size and $n\ll N$. Then, the optimization of the model parameters can be formulated by moving the current $w^t$ along the descent direction of training loss on a mini-batch training data:
\begin{equation}
 \widehat{w}^{t+1}(m_g^t)=w^t-\alpha \frac{1}{n}\sum_{j=1}^{n}\nabla_wL_j^T(w;m_{g_{j}}^t) \bigg|_{w=w^t} \label{model update}
\end{equation}
where $\alpha$ is the descent step size on $w$.

Then, a mini-batch of meta samples $\{(x_i^{v},y_i^{v}),1\leq i\leq n\}$ is sampled with $n/4$ samples per skin tone bin. We extract the features of these meta samples by the updated model parameters $ \widehat{w}^{t+1}(m_g^t)$, and calculate meta skewness loss $L^{M}$. The margin $m_g$ can then be readily updated guided by Eq. \ref{update margin}, i.e., moving the current parameter $m_g^t$ along the objective gradient of $L^{M}$ calculated on the meta-data:
\begin{equation}
{m_g}^{t+1}=m_g^t-\beta  \nabla_{m_g}L^M( \widehat{w}^{t+1}(m_g)) \bigg|_{m_g=m_g^t} \label{margin update}
\end{equation}
where $\beta$ is the descent step size on margin $m_g$. Note that we need to use meta loss to optimize the margins of training loss by computing the gradient of $L^{M}$ with respect to margin $m_g$. However, \emph{the skewness (bias) evaluated by meta skewness loss $L^{M}$ is incurred on meta data, while the margin $m_g$ only plays effect in adaptive margin loss $L^T$ in training phase}. We connect the disjoint process between training and meta-evaluating via the updated model parameter $ \widehat{w}^{t+1}(m_g^t)$. Notice that $\widehat{w}^{t+1}$ here is a variable instead of a quantity, which makes $ \widehat{w}^{t+1}(m_g^t)$ a function of $m_g^t$ and the gradient in Eq. \ref{margin update} be able to be computed. Thus, the meta-optimization (Eq. \ref{margin update}) can be performed over the margin parameter $m_g$ as follows:

\begin{footnotesize}
\begin{equation}
\begin{split}
{m_g}^{t+1}&={m_g}^t-\beta  \nabla_{m_g}L^M( \widehat{w}^{t+1}(m_g)) \bigg|_{m_g=m_g^t}\\
&=m_g^t-\beta  \frac{\partial L^M( \widehat{w}^{t+1}(m_g))}{\partial m_g} \bigg|_{m_g=m_g^t}\\
&=m_g^t-\beta  \frac{\partial L^M( \widehat{w})}{\partial \widehat{w}} \bigg|_{\widehat{w}=\widehat{w}^{t+1}} \frac{\partial  \widehat{w}^{t+1}(m_g)}{\partial m_g} \bigg|_{m_g=m_g^{t}}\\
&=m_g^t+\frac{\alpha \beta }{n} \sum_{j=1}^{n}\frac{\partial L^M( \widehat{w})}{\partial \widehat{w}} \bigg|_{\widehat{w}=\widehat{w}^{t+1}}\frac{\partial^2 L_j^T(w;m_{g_j})}{\partial w\partial m_g} \bigg|_{w=w^t,m_g=m_g^t} \label{meta gradient}
\end{split}
\end{equation}
\end{footnotesize}
Therefore, the meta-gradient update involves a gradient through a gradient, i.e., $\frac{\partial^2 L_j^T(w;m_{g_j})}{\partial w\partial m_g}$, which can transmit meta gradient from meta loss to training loss to update the margins. When implementing, we can leverage automatic differentiation techniques to compute it. We can first unroll the gradient graph of the training batch, and then use backward-on-backward automatic differentiation to take a second order gradient pass.

Then, the adaptive margin loss with updated margin $m_g^{t+1}$ is employed to ameliorate the model parameters $w$:
\begin{equation}
 w^{t+1}=w^t-\alpha \frac{1}{n}\sum_{j=1}^{n}\nabla_wL_j^T(w;m_{g_{j}}^{t+1}) \bigg|_{w=w^t} \label{model update2}
\end{equation}

The meta-learning algorithm can then be summarized in Algorithm \ref{al1} and Fig. \ref{iterative}, and Fig. \ref{MBN} illustrates its main implementation process. In our algorithm, both the model and margin gradually ameliorate their parameters during the learning process based on their values calculated in the last step. Finally, the optimal model parameters are learned which perform well and fairly across different skin tone bins.

\begin{algorithm}[htb]
\caption{ Meta balanced network.}
\label{al1}
\begin{algorithmic}[1]
\REQUIRE ~~\\
Training data $D_{train}$, meta data $D_{meta}$, batch size $n$ and max iterations $T$.
\ENSURE ~~\\
Model parameter $w$ and margin parameter $m_g$.
\STATE Initialize model parameter $w^0$ and margin parameter $m_g^0$.
\FOR {$t=0$ \textbf{to} $T-1$}
\STATE $\{X,Y\}\leftarrow$SampleMiniBatch$(D_{train}, n)$
\STATE $\{X^v,Y^v\}\leftarrow$SampleMiniBatch$(D_{meta}, n)$
\STATE $f(X,w^t)\leftarrow$Forward($X,w^t$). 
\STATE $L^T\leftarrow \frac{1}{n}\sum_{j=1}^{n}L_j^T(f(x_j;w^t);y_j;m_{g_{j}}^t)$ by Eq. \ref{RBN-arc} or \ref{RBN-cos}.
\STATE $\nabla w^t\leftarrow$BackwardAD$(L^T,w^t)$. 
\STATE $\widehat{w}^{t+1} \leftarrow w^t-\alpha \nabla w^t$ by Eq. \ref{model update}.
\STATE $f(X^v,\widehat{w}^{t+1})\leftarrow$Forward($X^v,\widehat{w}^{t+1}$). 
\STATE $L^M\leftarrow L^M(f(X^v;\widehat{w}^{t+1});Y^v)$ by Eq. \ref{meta loss1}, \ref{meta loss2} and \ref{diff_indian}.
\STATE $\nabla m_g^t\leftarrow$BackwardAD$(L^M,m_g^t)$. 
\STATE $m_g^{t+1} \leftarrow m_g^t-\beta \nabla m_g^t$ by Eq. \ref{meta gradient}.
\STATE $\widehat{L}^T\leftarrow \frac{1}{n}\sum_{j=1}^{n}L_j^T(f(x_j;w^t);y_j;m_{g_{j}}^{t+1})$ by Eq. \ref{RBN-arc} or \ref{RBN-cos}. 
\STATE $\nabla w^t\leftarrow$BackwardAD$(\widehat{L}^T,w^t)$. 
\STATE $w^{t+1} \leftarrow w^t-\alpha \nabla w^t$ by Eq. \ref{model update2}.
\ENDFOR
\end{algorithmic}
\end{algorithm}

\section{Experiments}

\subsection{Experimental study on bias} \label{section}

\begin{figure*}
\centering
\subfigure[Baidu]{
\label{fig5e} 
\includegraphics[width=3.8cm]{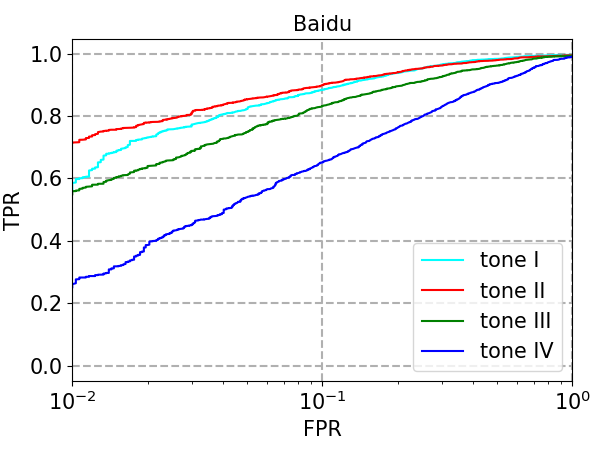}}
\hspace{0cm}
\subfigure[Face++]{
\label{fig0f} 
\includegraphics[width=3.8cm]{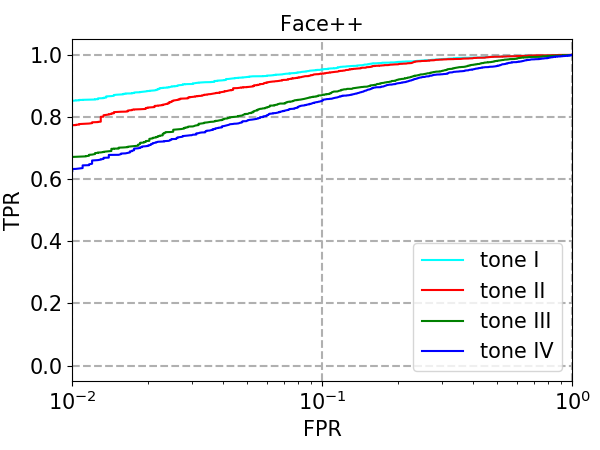}}
\hspace{0cm}
\subfigure[Amazon]{
\label{fig0g} 
\includegraphics[width=3.8cm]{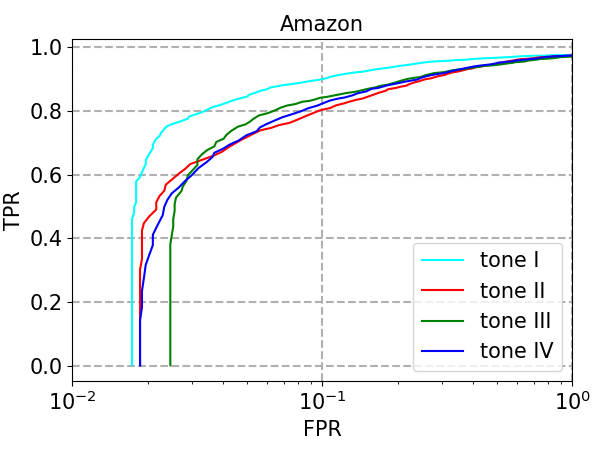}}
\hspace{0cm}
\subfigure[Microsoft]{
\label{fig0h} 
\includegraphics[width=3.8cm]{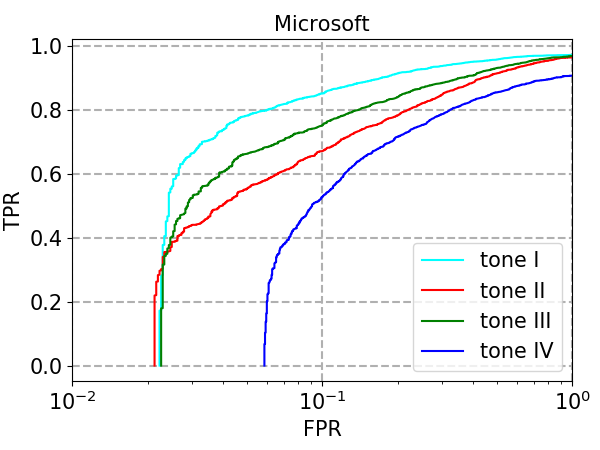}}
\caption{The ROC curves of (a) Baidu, (b) Face++, (c) Amazon and (d) Microsoft evaluated on 6K pairs of IDS-4. Due to limited number of negative pairs, the performances cannot be reliably estimated at lower FPR values. Besides, once API fails to detect faces, we assume that it will give an incorrect verification result whatever decision thresholds are.}
\label{fig5} 
\end{figure*}

\begin{figure*}
\centering
\subfigure[Baidu]{
\label{fig5e} 
\includegraphics[width=3.8cm]{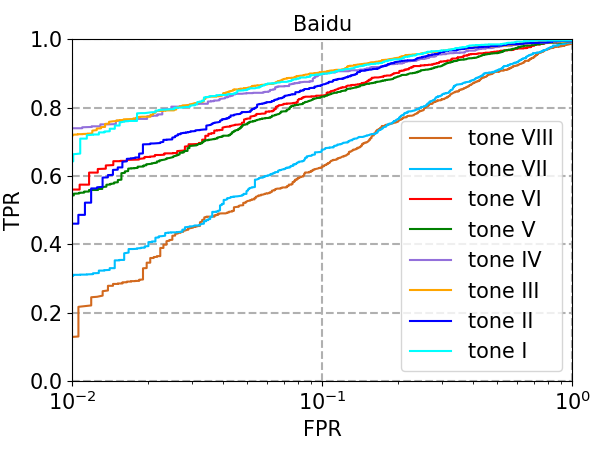}}
\hspace{0cm}
\subfigure[Face++]{
\label{fig0f} 
\includegraphics[width=3.8cm]{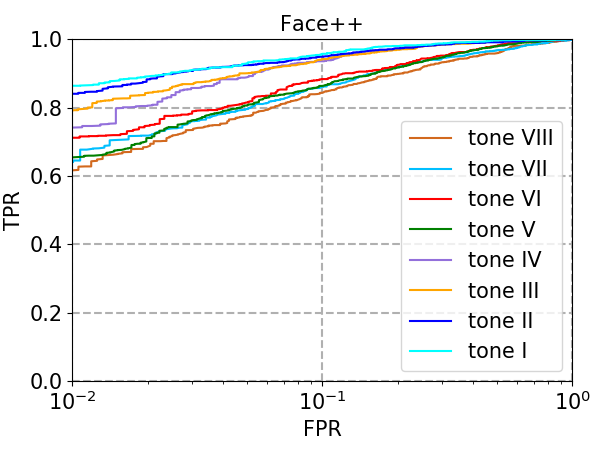}}
\hspace{0cm}
\subfigure[Amazon]{
\label{fig0g} 
\includegraphics[width=3.8cm]{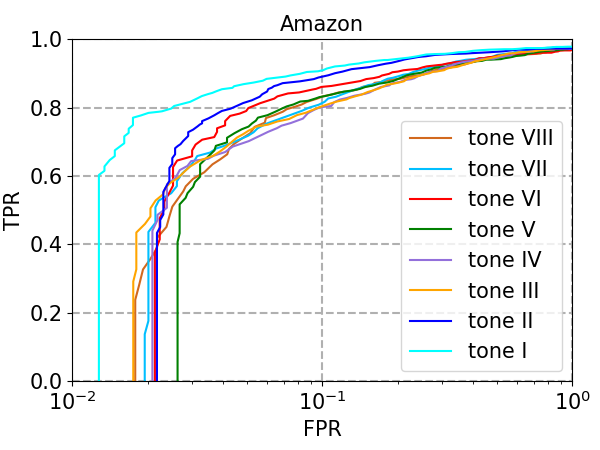}}
\hspace{0cm}
\subfigure[Microsoft]{
\label{fig0h} 
\includegraphics[width=3.8cm]{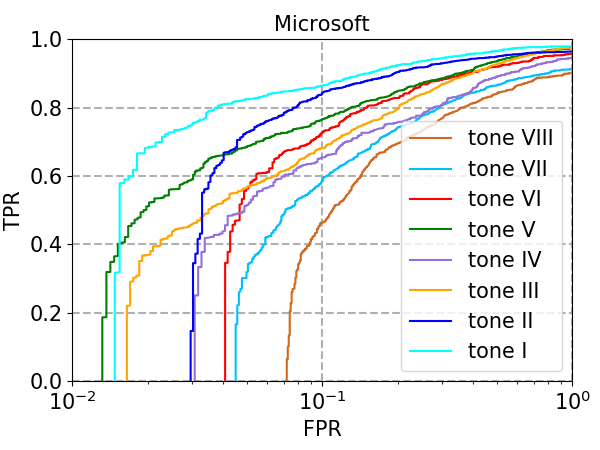}}
\caption{The ROC curves of (a) Baidu, (b) Face++, (c) Amazon and (d) Microsoft evaluated on 3K pairs of IDS-8. Due to limited number of negative pairs, the performances cannot be reliably estimated at lower FPR values. }
\label{fig5_8bin} 
\end{figure*}

\textbf{Experimental Settings.} We use the similar ResNet-34 architecture described in \cite{deng2018arcface}. It is trained with the guidance of Arcface loss \cite{deng2018arcface} on the CASIA-Webface \cite{yi2014learning}, and is called Arcface(CASIA) model. CASIA-Webface consists of 0.5M images of 10K celebrities in which 85\% of the photos are lighter-skinned subjects. For preprocessing, we use five facial landmarks for similarity transformation, then crop and resize the faces to 112$\times$112. Each pixel ([0, 255]) in RGB images is normalized by subtracting 127.5 and then being divided by 128. We set the batch size, momentum, and weight decay as 200, 0.9 and $5e-4$, respectively. LFW \cite{huang2007labeled}, CFP-FP \cite{sengupta2016frontal} and AgeDB-30 \cite{moschoglou2017agedb} are utilized as validations to determine when to decrease the learning rate or stop training. The learning rate is started from 0.1 and decreased twice with a factor of 10 when errors plateau on LFW, CFP-FP and AgeDB-30.

\begin{table}[htbp]
    \setlength{\abovecaptionskip}{0cm}
	\setlength{\belowcaptionskip}{-0.2cm}
    \footnotesize
    \caption{Bias with respect to skin tone in commercial APIs and SOTA FR algorithms. Accuracies (\%) on our IDS-4 are given. }
    \label{tab5}
	\begin{center}
    \begin{threeparttable}
    \setlength{\tabcolsep}{1.9mm}{
	\begin{tabular}{c|c|cccc}
		\hline
        \multirow{2}{*}{Model} &\multirow{2}{*}{LFW \cite{huang2007labeled}} & \multicolumn{4}{c}{IDS-4} \\
                               &                     & I & II & III & IV \\ \hline \hline
         Microsoft \cite{azure} &98.22 &\textcolor{red}{87.60} &\textcolor{red}{79.67} &\textcolor{red}{82.83} &\textcolor{red}{75.83}   \\
         Face++ \cite{Face++} &  \textcolor{red}{97.03}& \textbf{93.90} & \textbf{92.47}& \textbf{88.55}&  \textbf{87.50}\\
         Baidu \cite{Baidu} &\textbf{98.67} &89.13 & 90.27& 86.53& 77.97  \\
         Amazon \cite{amazon} &98.50 &90.45 &84.87 & 87.20 &86.27  \\ \hline
         mean & 98.11 & 90.27 & 86.82 & 86.28 &  81.89 \\ \hline
         Center-loss \cite{wen2016discriminative} &\textcolor{red}{98.75} & \textcolor{red}{87.18} & \textcolor{red}{79.32} &\textcolor{red}{81.92} & \textcolor{red}{78.00} \\
         Sphereface \cite{liu2017sphereface} &99.27 & 90.80 &82.95 & 87.02 &82.28  \\
         Arcface\tnote{1} \cite{deng2018arcface} & \textbf{99.40} & \textbf{92.15} &83.98 & \textbf{88.00} &\textbf{84.93}  \\
         VGGface2\tnote{2} \cite{cao2017vggface2} & 99.30 &89.90 &\textbf{84.93} & 86.13 & 83.38  \\ \hline
         mean & 99.18 & 90.01 & 82.80 & 85.77 &  82.15 \\ \hline
	\end{tabular}}
    \begin{tablenotes}
     \item[1] Arcface is a ResNet-34 model trained with CASIA-Webface. 
     \item[2] VGGFace2 here is a SeNet model trained with VGGFace2.
    \end{tablenotes}
    \end{threeparttable}
    \end{center}
\end{table}

\textbf{Existence of bias.} We examine some SOTA algorithms, i.e., Center-loss \cite{wen2016discriminative}, Sphereface \cite{liu2017sphereface}, VGGFace2 \cite{cao2017vggface2} and ArcFace \cite{deng2018arcface}, as well as four commercial recognition APIs, i.e., Face++, Baidu, Amazon, Microsoft on our IDS-4 and IDS-8, respectively.
The results on IDS-4 are presented in Table \ref{tab5}, Fig. \ref{fig5} and Fig.\ref{roc_SOTA}. All SOTA algorithms perform best on Tone-I bin and worst on subjects with skin tone II and IV, which proves the existence of bias with respect to skin tone. The results on IDS-8 are presented in Table \ref{racial bias of 8bin} and Fig. \ref{fig5_8bin}. The similar bias can be observed on IDS-8. For example, Arcface has an accuracy of 92.33\% for people with Tone-I but drops to 83.23\% for subjects with Tone-VIII.

\begin{figure}
\centering
\subfigure[t-SNE on IDS-4]{
\label{fig4a} 
\includegraphics[width=3.8cm]{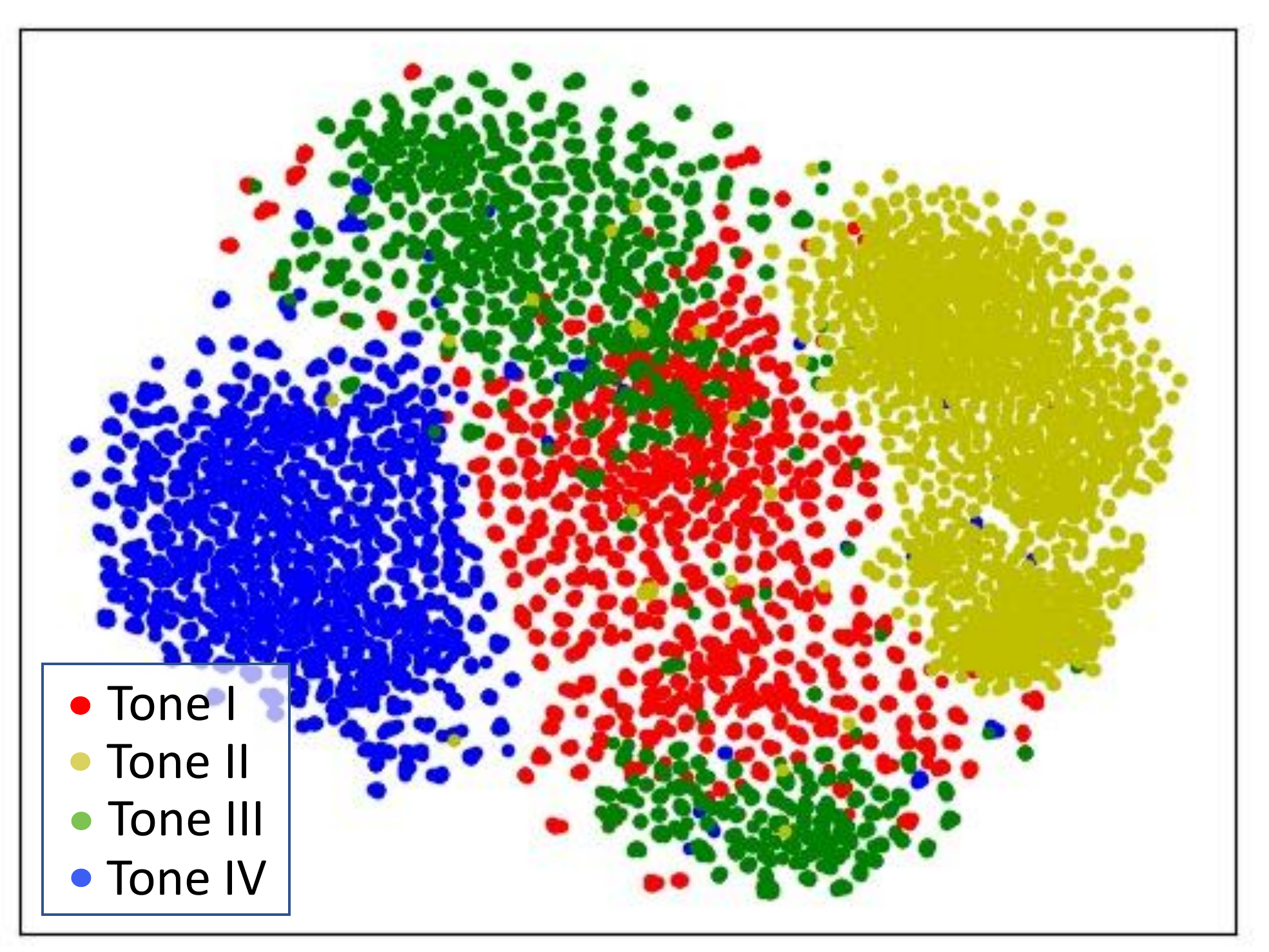}}
\hspace{0cm}
\subfigure[t-SNE on IDS-8]{
\label{fig4b} 
\includegraphics[width=3.8cm]{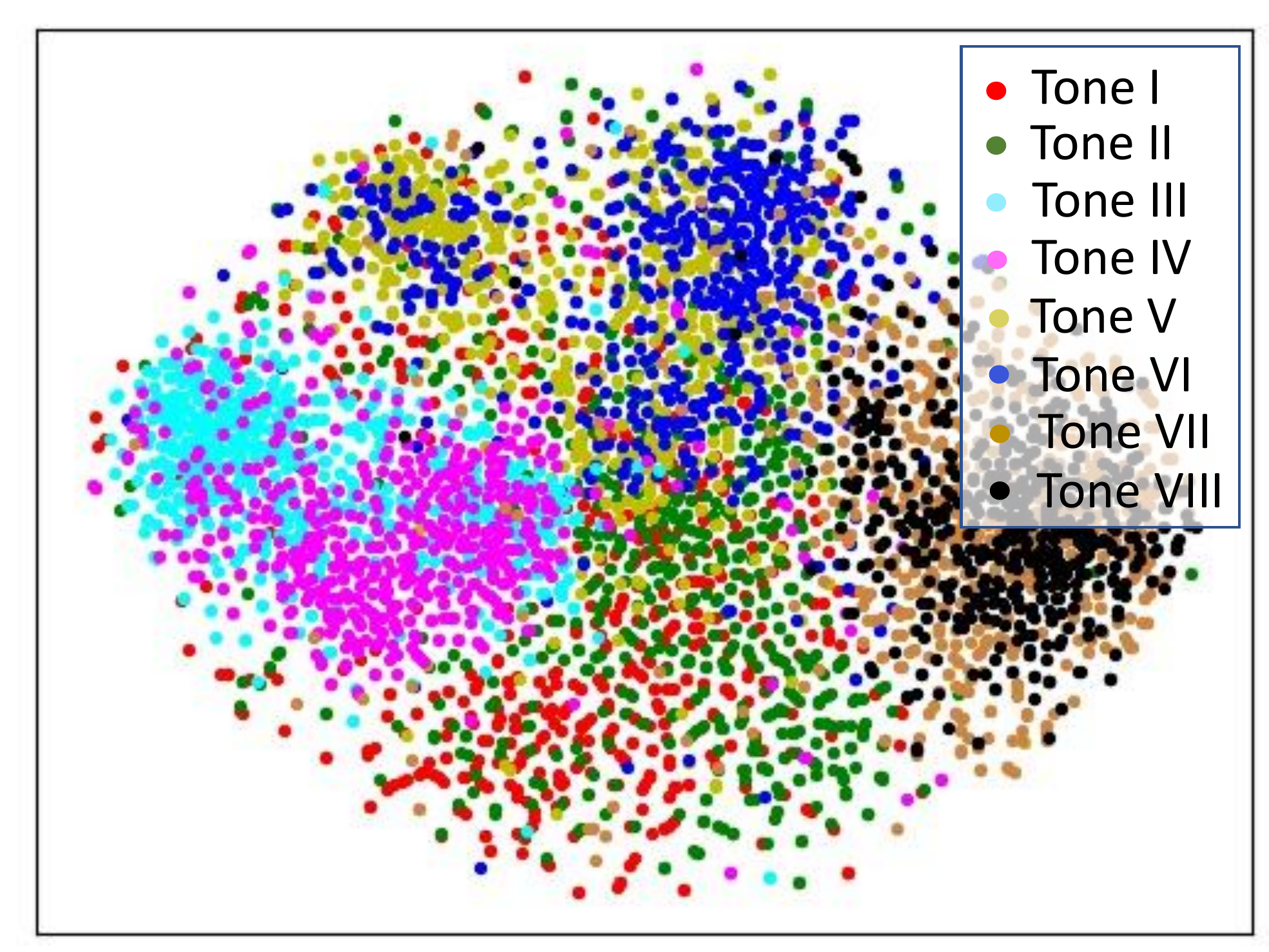}}
\caption{ The feature space of Arcface(CASIA) model.}
\label{fig4} 
\end{figure}

\begin{figure}
\centering
\subfigure[Center loss]{
\label{fig12a} 
\includegraphics[width=3.8cm]{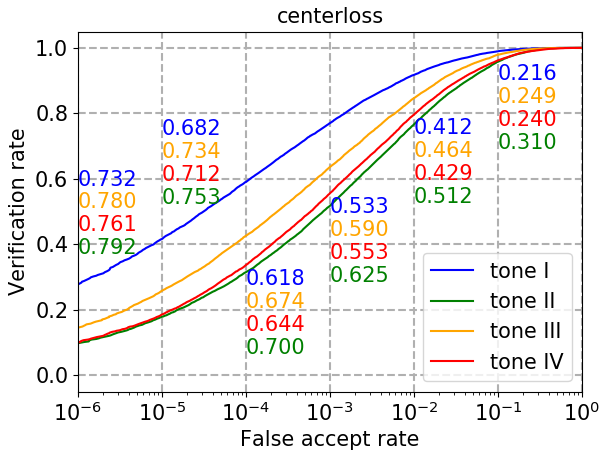}}
\subfigure[Spereface]{
\label{fig12b} 
\includegraphics[width=3.8cm]{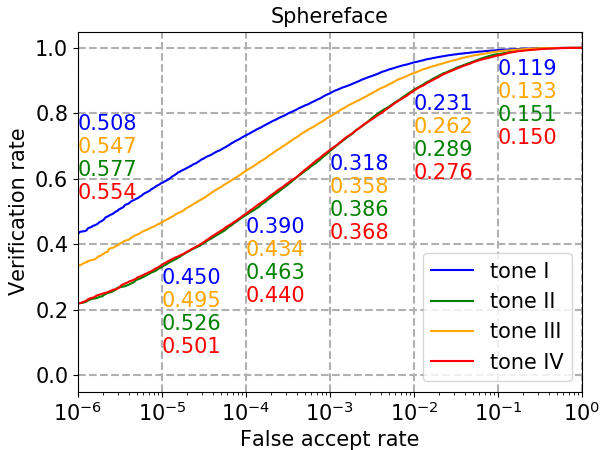}}
\subfigure[Arcface]{
\label{fig12c} 
\includegraphics[width=3.8cm]{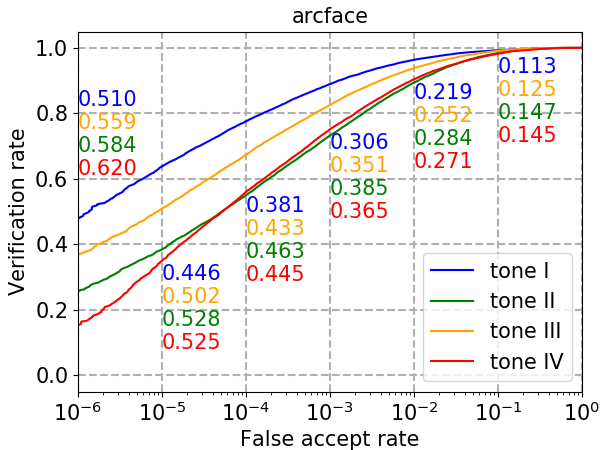}}
\subfigure[VGGFace2]{
\label{fig12d} 
\includegraphics[width=3.8cm]{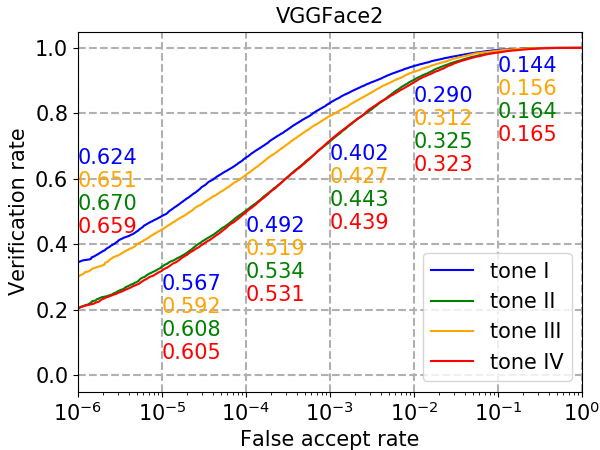}}
\caption{The ROC curves of (a) Center loss, (b) Spereface (c) Arcface, (d) VGGFace2 evaluated on all pairs of IDS-4. The cosine similarity thresholds of different skin tone bins are showed at each axis point (FAR=$\{10e-6,10e-5,10e-4,10e-3,10e-2,10e-1\}$).}
\label{roc_SOTA} 
\end{figure}

\begin{table*}[!htbp]
    \setlength{\abovecaptionskip}{0cm}
	\setlength{\belowcaptionskip}{-0.2cm}
    \caption{Verification accuracy (\%) of ResNet-34 models trained with CASIA-Webface \cite{yi2014learning} and our Balancedface$^{*}$.}
    \label{tab2}
    \footnotesize
	\begin{center}
    \setlength{\tabcolsep}{1.2mm}{
	\begin{tabular}{c|ccc|cccc|cccccccc}
		\hline
         \multirow{2}{*}{Training Databases} & \multirow{2}{*}{LFW \cite{huang2007labeled}} & \multirow{2}{*}{CFP-FP \cite{sengupta2016frontal}} & \multirow{2}{*}{AgeDB-30 \cite{moschoglou2017agedb}} & \multicolumn{4}{c|}{IDS-4} & \multicolumn{8}{c}{IDS-8}\\
         & & & & I & II & III & IV & I & II & III & IV & V & VI & VII & VIII \\ \hline \hline
         CASIA-WebFace \cite{yi2014learning} & 99.40 &\textbf{93.91} & 93.35 & 92.15 & 83.98 & 88.00 & 84.93 & 92.33 & 91.70 & 83.97 & 83.05 & 87.46 & 88.40 & 84.07 & 83.23\\ 	
         Balancedface$^{*}$ (ours)   & \textbf{99.55} & 92.74  &  \textbf{95.15} & \textbf{93.92} &  \textbf{90.60} & \textbf{92.98}	& \textbf{90.98} & \textbf{93.95} & \textbf{93.85} & \textbf{90.14} & \textbf{91.14} & \textbf{92.97} & \textbf{93.55} &   \textbf{91.70} & \textbf{90.32} \\ \hline
	\end{tabular}}
    \end{center}
\end{table*}

\textbf{Existence of domain gap.} We extract the features of 1.2K images by our Arcface(CASIA) model and visualize them using t-SNE embeddings \cite{donahue2014decaf} in Fig. \ref{fig4}. In Fig. \ref{fig4a}, although Arcface(CASIA) model is a face recognition model, not skin detection, the skin tone information is highly embedded in the feature space of IDS-4, resulting in four clusters highly correlated with skin tone. On IDS-8, because more fine-grained division according to skin tone increases the similarity of faces of different bins, there is not a clear boundary between these eight skin tone bins. However, they are also separated from each other as shown in Fig. \ref{fig4b}.
From these figures, we make the conclusions: the distribution discrepancies across skin tone bins are large, which conforms that there is domain gap between people with different skin tones.

\textbf{Cause of bias.} We further select a subset from BUPT-Balancedface, called Balancedface$^{*}$. It contains 590K images from 14K celebrities which has the similar scale with CASIA-Webface database but is approximately balanced with respect to skin tone. Using Balancedface$^{*}$ as training data, we train an Arcface(Balanced) model in the same way as Arcface(CASIA) model and compare their performances on IDS-4 and IDS-8, as shown in Table \ref{tab2}. Compared with Arcface(CASIA) model, Arcface(Balanced) model trained equally on all skin tones performs much better on darker-skinned subjects which proves that bias in databases will reflect in FR algorithm. However, even with balanced training, we see that darker-skinned subjects still perform poorly than subjects with skin tone I. The reason may be that faces of dark skin are inherently difficult to recognize. 

\begin{figure*}
\centering
\subfigure[Baidu]{
\label{fig5e} 
\includegraphics[width=3.8cm]{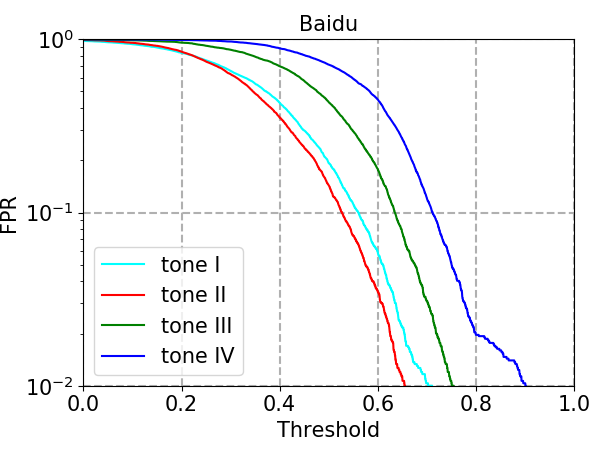}}
\hspace{0cm}
\subfigure[Face++]{
\label{fig0f} 
\includegraphics[width=3.8cm]{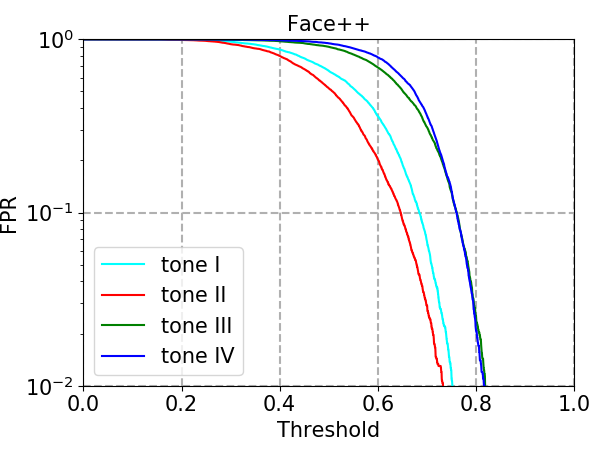}}
\hspace{0cm}
\subfigure[Amazon]{
\label{fig0g} 
\includegraphics[width=3.8cm]{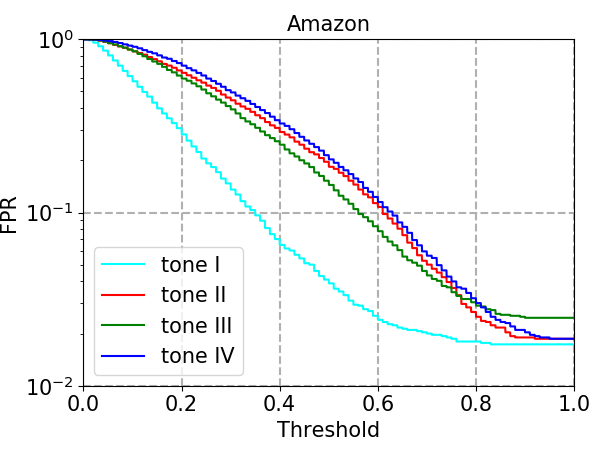}}
\hspace{0cm}
\subfigure[Microsoft]{
\label{fig0h} 
\includegraphics[width=3.8cm]{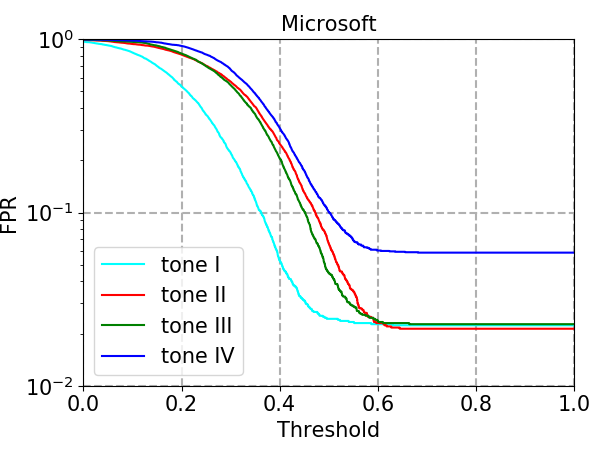}}
\caption{False positive rate is plotted as a function of threshold when (a) Baidu, (b) Face++, (c) Amazon and (d) Microsoft are evaluated on 6K pairs of IDS-4. Due to limited number of negative pairs, the performances cannot be reliably estimated at lower FPR values.}
\label{thre_4bin} 
\end{figure*}

\textbf{Decision thresholds of different skin tone groups.} FR algorithms use a threshold similarity score to determine whether two images are of the same subject. Different thresholds yield a different number of true/false positives and true/false negatives, and consequently different accuracy for a given dataset. Here, we look at the influences of the choice of threshold cutoff on performances of different skin tone bins. In Fig. \ref{thre_4bin}, we plot the false positive rate (FPR) as a function of threshold when evaluating commercial APIs on IDS-4. For ROC curves of SOTA algorithms, thresholds are also showed at each axis point in Fig. \ref{roc_SOTA}. In all cases, the thresholds that produce the same FPRs are shifted for different bins showing that threshold is skin-tone-specific, i.e., different for different skin tone groups. Therefore, to operate at a particular FPR, threshold should be carefully selected taking skin tone into consideration.

\begin{figure*}[!htbp]
\centering
\subfigure[Illumination (accuracy)]{
\label{v-light1} 
\includegraphics[width=4.15cm]{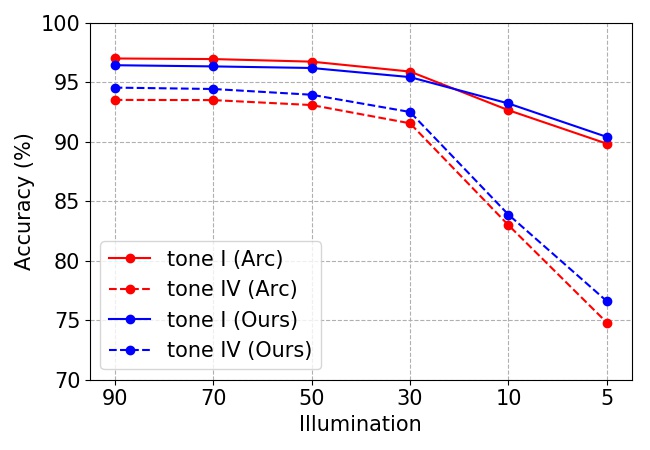}}
\hspace{0cm}
\subfigure[Illumination (skewness)]{
\label{v-light2} 
\includegraphics[width=4.15cm]{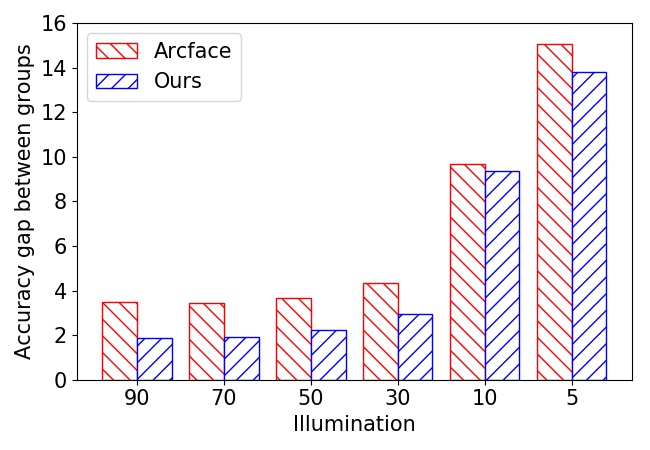}}
\subfigure[Occlusion (accuracy)]{
\label{v-occlu1} 
\includegraphics[width=4.15cm]{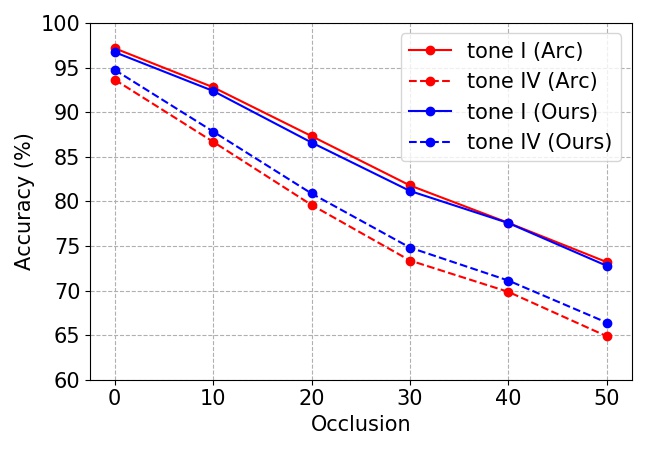}}
\hspace{0cm}
\subfigure[Occlusion (skewness)]{
\label{v-occlu2} 
\includegraphics[width=4.15cm]{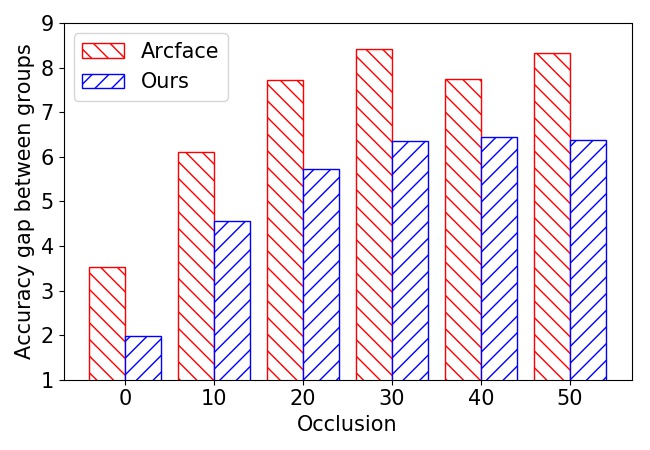}}
\hspace{0cm}
\subfigure[Noise (accuracy)]{
\label{v-noise1} 
\includegraphics[width=4.15cm]{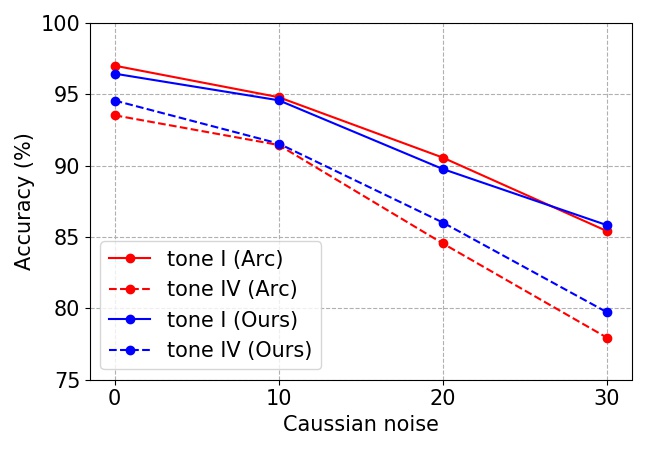}}
\hspace{0cm}
\subfigure[Noise (skewness)]{
\label{v-noise2} 
\includegraphics[width=4.15cm]{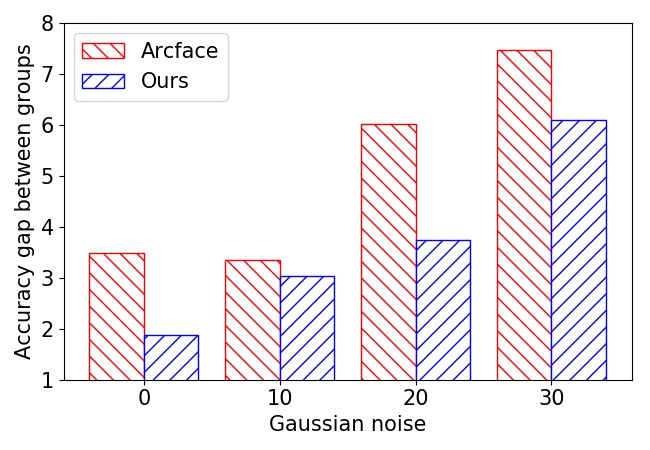}}
\hspace{0cm}
\subfigure[Pose (accuracy)]{
\label{v-pose1} 
\includegraphics[width=4.15cm]{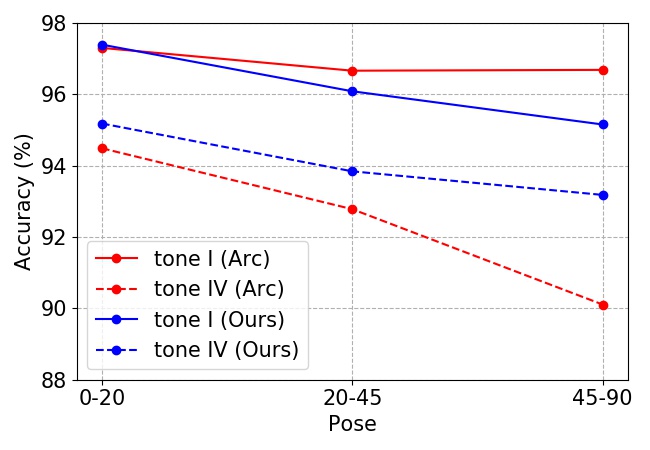}}
\hspace{0cm}
\subfigure[Pose (skewness)]{
\label{v-pose2} 
\includegraphics[width=4.15cm]{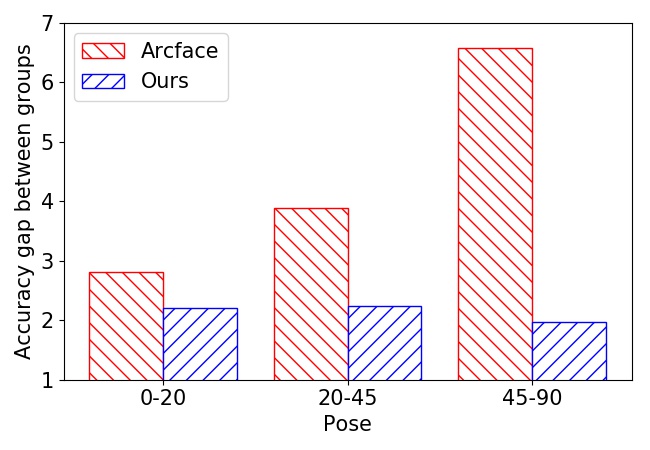}}
\hspace{0cm}
\subfigure[Age (accuracy)]{
\label{v-age1} 
\includegraphics[width=4.15cm]{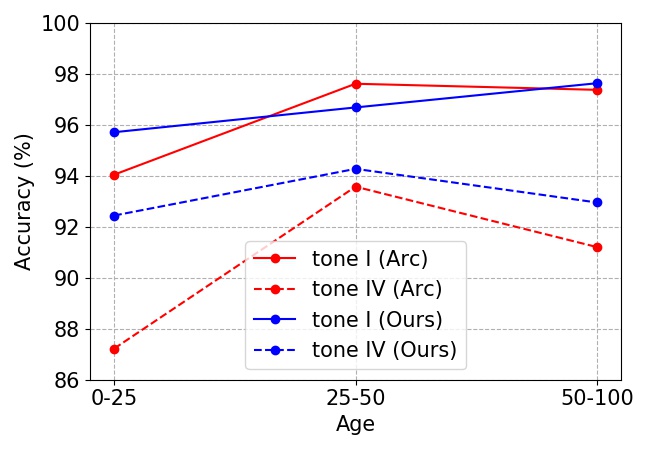}}
\hspace{0cm}
\subfigure[Age (skewness)]{
\label{v-age2} 
\includegraphics[width=4.15cm]{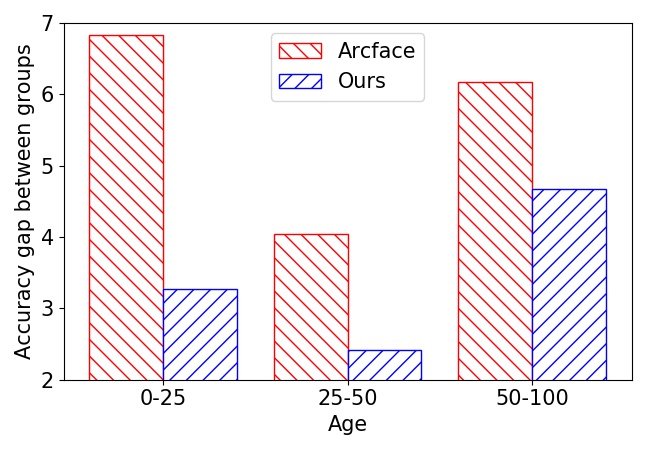}}
\hspace{0cm}
\subfigure[Gender (accuracy)]{
\label{v-gender1} 
\includegraphics[width=4.15cm]{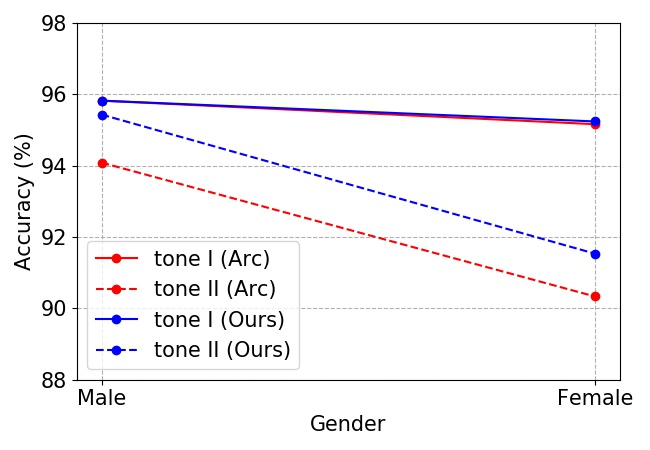}}
\hspace{0cm}
\subfigure[Gender (skewness)]{
\label{v-gender2} 
\includegraphics[width=4.15cm]{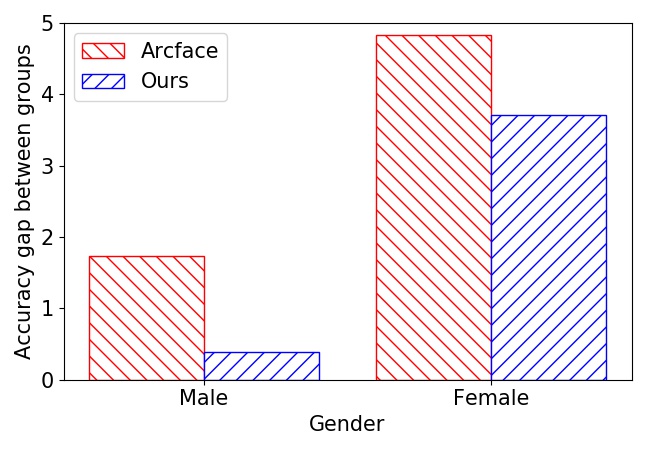}}
\caption{The influence of different variations on accuracy and skewness (accuracy gap between different skin tone groups) of Arcface and our MBN model. Lower accuracy gap is better and means fairer performance across groups.}
\label{variation} 
\end{figure*}

\textbf{Correlation of bias with other variations.} First, we degrade images of IDS-4 by occlusion, low illumination and Gaussian noise, and observe the influence of these variations on accuracy gap between Tone-I and Tone-IV subjects. We train ResNet-34 with guidance of Arcface loss \cite{deng2018arcface} using Globalface-4 dataset, and show the results using red lines in Fig. \ref{variation}. We can see that subjects with skin tone I and IV are both found to be sensitive to illumination, occlusion and Gaussian noise. Increasing variation level decreases the accuracy and fairness dramatically. Second, the correlations of bias with pose, age and gender are also studied by cohort analysis within each skin tone bin. For pose, we partition images of each skin tone bin into cohorts of (1) small-pose (0$^{\circ}$-20$^{\circ}$), (2) middle-pose (20$^{\circ}$-45$^{\circ}$) and (3) large-pose (45$^{\circ}$-90$^{\circ}$). For age, we partition the images into three cohorts: (1) young (0 to 25 years old), (2) middle-age (25 to 50 years old), and (3) old (50 to 100 years old). For gender, we partition the images into cohorts of (1) male and (2) female. We evaluate the trained Arcface model on different pose and age cohorts of Tone-I and Tone-IV bins. Because females with skin tone IV are too few to be evaluated, we evaluate the model on gender cohorts of our Tone-I and Tone-II bins instead. As we can see from red lines in Fig. \ref{variation}(g)-(l), for Arcface model, fairness with respect to skin tone decreases on large pose, young and female cohorts. Therefore, we conclude that faces with darker skin are more susceptible to different variations than the lighter-skinned ones, which results in a higher bias across groups when more difficult faces with large variations are recognized. 

\subsection{Meta margin learning experiment}

In this section, we verify the effectiveness of our MBN.

\textbf{Datasets.} We use our Globalface and Balancedface datasets to train our models, and use IDS to fairly measure performances of different skin tone groups. Moreover, in order to adopt meta learning to learn adaptive margins, we additionally collect a meta set with clean labels and balanced distribution. It contains 500 identities per skin tone bin and has been carefully and manually cleaned. We further remove its overlapping subjects with our training and testing datasets.

\begin{table*}[h]
    \setlength{\abovecaptionskip}{0cm}
	\setlength{\belowcaptionskip}{-0.2cm}
    \caption{Results on the verification experiments by varying distribution in the training set. Fairness is measured by the standard deviation (STD) (lower is better) and the skewed error ratio (SER) (1 is the best).}
    \label{simulate}
	\begin{center}
	\footnotesize
	\begin{tabular}{cc|cccc|c|cc}
		\hline
         \multicolumn{2}{c|}{Test$\rightarrow$} & \multicolumn{4}{c|}{IDS-4} & \multirow{2}{*}{Avg} & \multicolumn{2}{c}{Fairness} \\
         Train ratio $\downarrow$ & Method$\downarrow$ & I & II & III & IV & & STD & SER  \\ \hline \hline
         \multirow{2}{*}{$4:2:2:2$}&N-Softmax \cite{wang2017normface} & 89.67 & 84.68 & 87.97 &  84.17 & 86.62 & 2.64 & 1.53 \\
         &MBN(soft)& 91.05 & 88.10 & 90.17  & 88.80 & 89.53 & \textbf{1.33} & \textbf{1.32} \\ \hline
         \multirow{2}{*}{$5:\frac{5}{3}:\frac{5}{3}:\frac{5}{3}$}&N-Softmax \cite{wang2017normface} & 89.88 & 85.13 & 88.52 &  83.42 & 86.74 & 2.98 & 1.64 \\
         &MBN(soft)&91.28 & 87.73 & 90.68 &  88.15 & 89.46 & \textbf{1.78} & \textbf{1.41} \\ \hline
         \multirow{2}{*}{$6:\frac{4}{3}:\frac{4}{3}:\frac{4}{3}$}&N-Softmax \cite{wang2017normface} & 90.43 & 84.75 & 88.32 &  83.32 & 86.70 & 3.26 & 1.74 \\
         &MBN(soft)& 91.28 & 87.57 & 90.35 &  87.22 & 89.10 & \textbf{2.02} & \textbf{1.47} \\ \hline
         \multirow{2}{*}{$7:1:1:1$}&N-Softmax \cite{wang2017normface} & 90.67 & 84.37 & 87.77 & 82.97 & 86.44 & 3.46 & 1.83 \\
         &MBN(soft)& 90.85 & 86.82 & 89.20 &  86.08 & 88.24 & \textbf{2.19} & \textbf{1.52} \\ \hline
	\end{tabular}
    \end{center}
\end{table*}

\textbf{Experimental Settings.} For preprocessing, we share the uniform alignment methods and backbone as Arcface(CASIA) model as mentioned above. The batch size is set to be 240. LFW \cite{huang2007labeled}, CFP-FP \cite{sengupta2016frontal} and AgeDB-30 \cite{moschoglou2017agedb} are utilized as validations to determine when to decrease the learning rate or stop training. For model optimization, the learning rate $\alpha$ starts from 0.1 and is divided by 10 at 80K, 120K, 155K iterations on Globalface. The training process is finished at 180K iterations when errors reach a plateau on LFW, CFP-FP and AgeDB-30. On Balancedface, we divide the learning rate at 70K, 100K, 120K iterations and finish at 145K iterations. 

We train the models with SGD and set momentum as 0.9 and weight decay as $5e-4$. For meta-optimization, we use SGD to optimize the margin parameters with the momentum of 0.9. The learning rate $\beta$ begins with $1e-3$ and is decreased twice with a factor of 10. 
The hyper-parameter $\gamma$ in meta skewness loss $L^M$ is 0.5. When computing $L_M$, the triplets (anchor, positive and negative samples) are selected online during the training process for efficiency. Among all the triplets in the generated batches, the online selection chooses those for which: $\left \| f(x_{i}^{v})-f(x_{i,n}^{v}) \right \| _2^{2}-\left \| f(x_{i}^{v})-f(x_{i,p}^{v}) \right \|_2^{2}\leq 0.2$.

When training models on datasets with 4 skin bins, we utilize Tone-I subjects as anchors, and keep their margins unchanged. Optimal margins are learned for other skin tone groups to mitigate their performance biases compared with anchors. According to the loss function and margin of anchors, three variations of our MBN are defined. In MBN(soft), Norm-Softmax \cite{wang2017normface} is adopted by Tone-I subjects; while MBN(arc) utilizes Arcface loss \cite{deng2018arcface} to optimize anchors and their margins are set to be 0.3. In these two above cases, other skin tone groups are optimized by Arcface loss and their margins are initialized by 0.3. In MBN(cos), Cosface loss is used. Margins are set to be 0.15 for people with skin tone I and are initialized by 0.15 for other groups. When training models on datasets with 8 skin bins, we utilize faces with skin tone I and II as anchors and similar experimental settings are taken.

For evaluating, we test deep models on our IDS dataset and report their accuracies. Following the metric proposed in \cite{wang2019mitigate}, we utilize average accuracy of different skin tone bins to evaluate total recognition performance, and use the standard deviation (STD) and the skewed error ratio (SER) to evaluate the fairness performance. STD reflects the amount of dispersion of accuracies of different bins. 
SER is calculated as the highest error rate divided by the lowest one across skin tone bins, which is formulated as $SER=\frac{\mathop{max}\limits_{g}Error_{g}}{\mathop{min}\limits_{g}Error_{g}}$ and $g$ means skin tone bin. 

\textbf{Results on simulated dataset.} We follow the same setting of \cite{wang2019mitigate} and evaluate the effectiveness of MBN on some simulated training sets with different skin tone distributions. The images of each training set are selected from BUPT-Globalface-4 and have the similar scale with CASIA-Webface \cite{yi2014learning}. The ratio of Tone-I subjects varies from $\frac{2}{5}$ to $\frac{7}{10}$. 
Norm-Softmax \cite{wang2017normface}, which normalizes weights and features based on Softmax, is compared with our MBN.

The results of IDS-4 reported in Table \ref{simulate} exhibit large gaps between the performances of different skin tone groups, suggesting that training on biased datasets results in algorithmic bias. 
Second, the skin tone distribution of the training set generally has a clear impact on the performances of different groups. With the change of distribution (from 4:2:2:2 to 7:1:1:1), we observe a decrease in fairness between different bins, indicative of more uneven distribution with greater bias. Third, our MBN(soft) significantly improves performance on darker-skinned (II-IV) subjects and obtains more balanced performance than Norm-Softmax on different skin tone groups.  
It verifies the effectiveness of the idea of adaptive margin and the superiority of MBN(soft) under different skin tone distributions.

\begin{figure}
\centering
\subfigure[Softmax]{
\label{balance_roc1} 
\includegraphics[width=3.8cm]{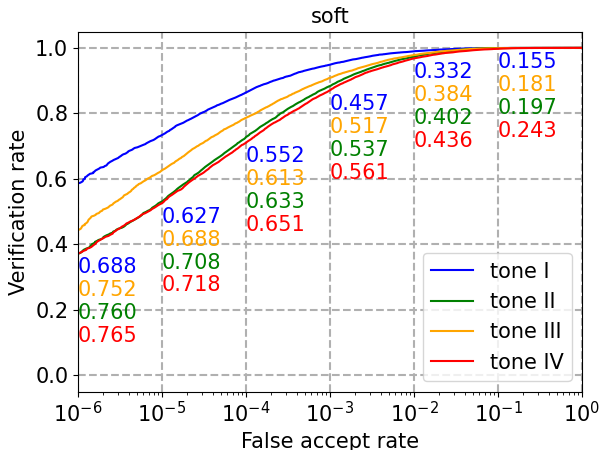}}
\subfigure[MBN(soft)]{
\label{balance_roc2} 
\includegraphics[width=3.8cm]{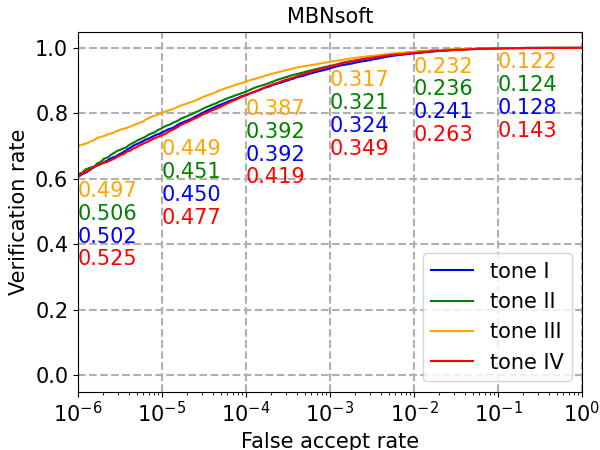}}
\subfigure[Cosface]{
\label{balance_roc3} 
\includegraphics[width=3.8cm]{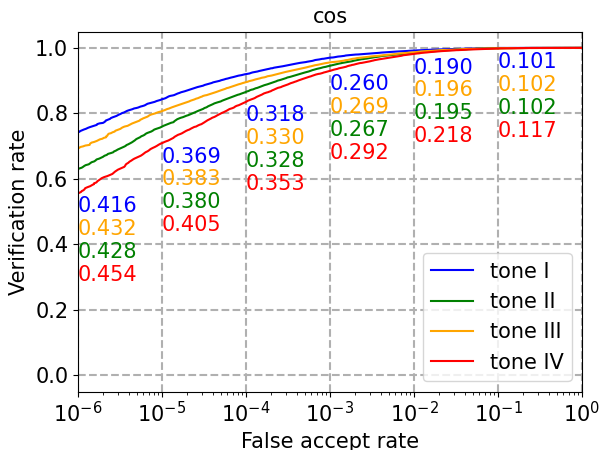}}
\subfigure[MBN(cos)]{
\label{balance_roc4} 
\includegraphics[width=3.8cm]{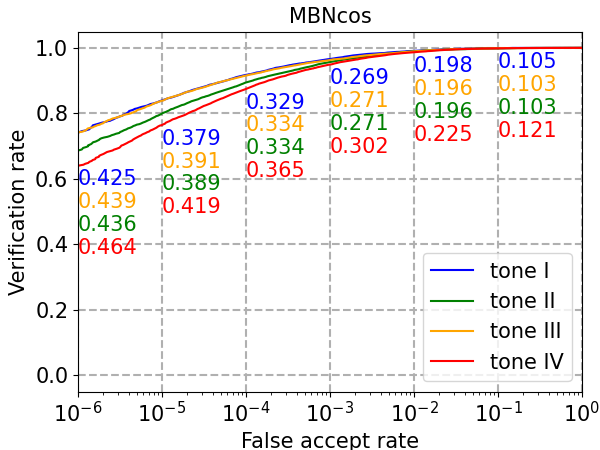}}
\subfigure[Arcface]{
\label{balance_roc5} 
\includegraphics[width=3.8cm]{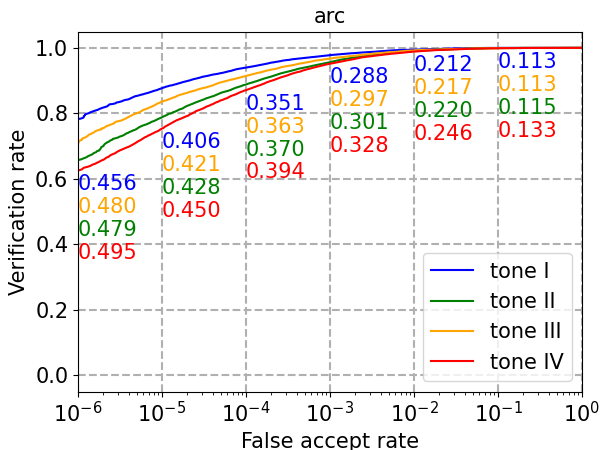}}
\subfigure[MBN(arc)]{
\label{balance_roc6} 
\includegraphics[width=3.8cm]{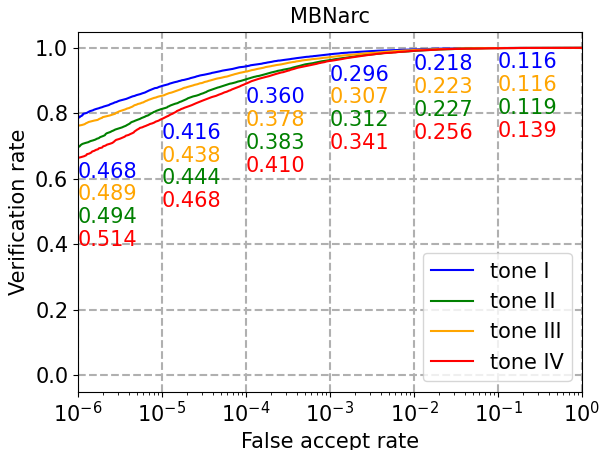}}
\caption{The ROC curves of (a) Softmax, (b) MBN(soft) (c) Cosface \cite{wang2018cosface}, (d) MBN(cos), (e) Arcface \cite{deng2018arcface} and (f) MBN(arc) evaluated on all pairs of IDS-4. The cosine similarity thresholds of different bins are showed at each axis point (FAR=$\{10e-6,10e-5,10e-4,10e-3,10e-2,10e-1\}$).}
\label{balance_roc} 
\end{figure}

\begin{table}[htbp]
    \setlength{\abovecaptionskip}{0cm}
	\setlength{\belowcaptionskip}{-0.2cm}
    \caption{Verification accuracy (\%) of methods trained with different loss function ([BUPT-Globalface-4, ResNet34, loss*]).}
    \label{worldface}
	\begin{center}
	\footnotesize
    \setlength{\tabcolsep}{1.0mm}{
	\begin{tabular}{c|cccc|c|cc}
		\hline
         \multirow{2}{*}{Methods} & \multicolumn{4}{c|}{IDS-4}  & \multirow{2}{*}{Avg} & \multicolumn{2}{c}{Fairness} \\
         & I & II & III & IV & & STD & SER \\ \hline \hline		
         Fisher vector \cite{simonyan2013fisher} & 70.23 & 66.88 & 69.55 &  63.22 & 67.47 & 3.18 & 1.24 \\
         Triplet \cite{schroff2015facenet} & 95.80 & 91.03 & 92.77 &  90.47 & 92.52 & 2.40 & 2.27 \\ \hline\hline
         Softmax & 95.62 & 90.85 & 91.97 &  89.98 &  \textcolor{red}{92.10} & \textcolor{red}{2.48} & \textcolor{red}{2.29} \\
         M-RBN(soft) \cite{wang2019mitigate} & 93.50 &  90.06 & 94.50 & 93.43 & 92.87 & 1.94 & 1.81 \\
         RL-RBN(soft) \cite{wang2019mitigate} & 94.53 & 94.20 & 95.03 &  94.05 & \textbf{94.45} & 0.44 & 1.20 \\
         MBN(soft)& 94.62 & 94.18 & 94.72 &  93.87 & 94.35 &  \textbf{0.39} &  \textbf{1.16} \\ \hline
         Cosface \cite{wang2018cosface} & 96.63 & 93.50 & 94.68 &  92.17 & \textcolor{red}{94.25} & \textcolor{red}{1.90} & \textcolor{red}{2.33} \\
         M-RBN(cos) \cite{wang2019mitigate} & 96.15 & 93.43 & 95.73 &  94.76 & 95.02 & 1.21 & 1.70 \\
         RL-RBN(cos) \cite{wang2019mitigate} & 96.03 & 94.58 & 95.15 &  94.27 & 95.01 & 0.77 & 1.45 \\
        	 MBN(cos)& 96.07 & 94.87 & 95.52 &  94.43 & \textbf{95.22} & \textbf{0.72} & \textbf{1.42} \\ \hline
         Arcface \cite{deng2018arcface} & 97.37 & 94.55 & 95.68 &  93.87 & \textcolor{red}{95.37} & \textcolor{red}{1.53} & \textcolor{red}{2.33} \\
         M-RBN(arc) \cite{wang2019mitigate} & 97.03 & 94.40 & 95.58 &  95.18 & 95.55 & 1.10 & 1.89 \\
         RL-RBN(arc) \cite{wang2019mitigate} & 97.08 & 95.57 & 95.63 &  94.87 & 95.79 & 0.93 & 1.76 \\
         MBN(arc)& 96.87 & 95.63 & 96.20 &  95.00 & \textbf{95.93} & \textbf{0.80} &  \textbf{1.59} \\ \hline 	
	\end{tabular}}
    \end{center}
\end{table}

\begin{table*}[htbp]
    \setlength{\abovecaptionskip}{0cm}
	\setlength{\belowcaptionskip}{-0.2cm}
    \caption{Verification accuracy (\%) of methods trained with different loss function ([BUPT-Globalface-8, ResNet34, loss*]).}
    \label{worldface-8bin}
	\begin{center}
	\footnotesize
    \setlength{\tabcolsep}{2mm}{
	\begin{tabular}{c|cccccccc|c|cc}
		\hline
         \multirow{2}{*}{Methods} & \multicolumn{8}{c|}{The skin tone of IDS-8} & \multirow{2}{*}{Avg} & \multicolumn{2}{c}{Fairness} \\
         & I & II & III & IV & V & VI & VII & VIII & & STD & SER \\ \hline \hline		
         Softmax & 95.06 & 94.98 & 90.29 & 90.02 & 90.64 & 93.02 & 89.88 & 87.97 & 91.48 & 2.58 & 2.43 \\
         MBN(soft)& 94.05 & 94.18 & 93.47 & 93.75 & 93.37 & 94.94 & 92.96 & 92.34 & \textbf{93.63} & \textbf{0.80} &  \textbf{1.51} \\ \hline
         Cosface \cite{wang2018cosface} & 96.30 & 95.97 & 93.19 & 93.08 & 94.16 & 94.56 & 92.86 & 91.57 & 93.96 & 1.61 & 2.28 \\
        	 MBN(cos)& 95.87 & 95.84 & 94.64 & 94.82 & 94.70 & 96.20 & 94.58 & 94.12 & \textbf{95.09} & \textbf{0.76} & \textbf{1.55} \\ \hline
         Arcface \cite{deng2018arcface} & 97.34 & 97.09 &  94.43 & 94.72 & 95.36 & 95.86 & 94.25 & 92.78 & 95.23 & 1.52 & 2.72 \\
         MBN(arc)& 97.11 & 96.83 &  94.94 & 95.45 & 95.54 & 96.73 & 94.81 & 94.83 & \textbf{95.78} & \textbf{0.96} & \textbf{1.79} \\ \hline 	
	\end{tabular}}
    \end{center}
\end{table*}

\textbf{Results on BUPT-Globalface dataset.} We utilize our BUPT-Globalface dataset to validate the effectiveness of our method on large-scale training data. 
We compare our MBN with Softmax, Cosface \cite{wang2018cosface} and Arcface \cite{deng2018arcface}. Cosface \cite{wang2018cosface} and Arcface \cite{deng2018arcface} assign the same margin for all images and report SOTA performance on the LFW \cite{huang2007labeled} and MegaFace \cite{kemelmacher2016megaface}. The scaling parameter is set as 60 for Cosface \cite{wang2018cosface} and Arcface \cite{deng2018arcface}; the margin parameters are set as 0.2 and 0.3, respectively. We train the models on Globalface-4 and show the results tested on IDS-4 in Table \ref{worldface} and Fig. \ref{balance_roc}. 
First, with the help of more separate inter-class data, Cosface \cite{wang2018cosface} and Arcface \cite{deng2018arcface} outperform Softmax by improving the fairness metrics. However, bias cannot be eliminated completely by a uniform margin for different skin tone groups. The performance of darker-skinned subjects is still inferior to that of people with skin tone I. Second, MBN(soft), MBN(cos) and MBN(arc) obtain fairer performances than Softmax, Cosface and Arcface. For example, MBN(arc) is superior to Arcface by reducing the SER from 2.33 to 1.59. It shows the effectiveness of our algorithm on learning balanced features from a biased dataset. Moreover, we additionally train the models on Globalface-8 and show the results tested on IDS-8 in Table \ref{worldface-8bin}. The same conclusion can be observed when the images are divided into 8 skin bins. Cosface and Arcface still show biased performance on faces with different skin tones, and our method performs more fairly benefitting from adaptive margin. For example, MBN(cos) is superior to Cosface by reducing the SER from 2.28 to 1.55 on IDS-8.

\begin{table}[htbp]
    \setlength{\abovecaptionskip}{0cm}
	\setlength{\belowcaptionskip}{-0.2cm}
    \caption{Verification accuracy (\%) of methods trained with different loss function ([BUPT-Balancedface-4, ResNet34, loss*]).}
    \label{balancedface}
	\begin{center}
	\footnotesize
        \setlength{\tabcolsep}{1.0mm}{
	\begin{tabular}{c|cccc|c|cc}
		\hline
         \multirow{2}{*}{Methods} & \multicolumn{4}{c|}{IDS-4}  & \multirow{2}{*}{Avg} & \multicolumn{2}{c}{Fairness} \\
         & I & II & III & IV & & STD & SER \\ \hline \hline
         Fisher vector \cite{simonyan2013fisher} & 70.23 & 66.88 & 69.55 &  63.22 & 67.47 & 3.18 & 1.24 \\
         Triplet \cite{schroff2015facenet} & 94.58 & 91.48 & 93.17 &  91.60 & 92.71 & 1.47 & 1.57 \\ \hline\hline
         Softmax & 94.18 & 91.23 & 92.82 &  91.42 & \textcolor{red}{92.37} & \textcolor{red}{1.42} & \textcolor{red}{1.51} \\
         RL-RBN(soft) \cite{wang2019mitigate} & 94.30 & 93.87 & 94.13 &  94.45 &  \textbf{94.19} &  0.25 &  1.10 \\
         MBN(soft)& 93.45 & 93.82 & 93.90 &  93.83 & 93.75 & \textbf{0.20} & \textbf{1.07} \\ \hline
         Cosface \cite{wang2018cosface} & 95.12 & 92.98 & 93.93 &  92.93 &  \textcolor{red}{93.74} &  \textcolor{red}{1.03} &  \textcolor{red}{1.45} \\
         RL-RBN(cos) \cite{wang2019mitigate} & 95.47 &94.52 &  95.15 &  95.27 & \textbf{95.10} & 0.41 & 1.21 \\
         MBN(cos)& 95.37 & 94.43 & 95.17 &  95.05 & 94.99 & \textbf{0.39} & \textbf{1.19} \\ \hline
         Arcface \cite{deng2018arcface} & 96.18 & 93.72 & 94.67 &  93.98 & \textcolor{red}{94.64} & \textcolor{red}{1.11} & \textcolor{red}{1.65} \\
         RL-RBN(arc) \cite{wang2019mitigate} & 96.27 & 94.82 & 94.68 &  95.00 & 95.19 & 0.73 & 1.43 \\
         MBN(arc)& 96.25 & 94.85 & 95.32 &  95.38 & \textbf{95.45} & \textbf{0.58} & \textbf{1.37} \\ \hline 	
	\end{tabular}}
    \end{center}
\end{table}

\begin{table*}[htbp]
    \setlength{\abovecaptionskip}{0cm}
	\setlength{\belowcaptionskip}{-0.2cm}
    \caption{Verification accuracy (\%) of methods trained with different loss function ([BUPT-Balancedface-8, ResNet34, loss*]).}
    \label{balancedface-8bin}
	\begin{center}
	\footnotesize
    \setlength{\tabcolsep}{2mm}{
	\begin{tabular}{c|cccccccc|c|cc}
		\hline
         \multirow{2}{*}{Methods} & \multicolumn{8}{c|}{The skin tone of IDS-8} & \multirow{2}{*}{Avg} & \multicolumn{2}{c}{Fairness} \\
         & I & II & III & IV & V & VI & VII & VIII & & STD & SER \\ \hline \hline		
         Softmax & 94.28 & 93.09 & 90.70 & 90.75 & 91.49 & 93.31 &  91.27 & 91.57 & 92.06 & 1.33 & 1.63 \\
         MBN(soft)& 93.61 & 93.23 & 93.21 & 92.78 & 92.89 & 94.55 &  93.82 & 93.11& \textbf{93.40} &  \textbf{0.58} &  \textbf{1.33} \\ \hline
         Cosface \cite{wang2018cosface} & 94.38 & 94.58 & 92.07 & 92.30 & 93.68 & 94.70 & 92.99 & 91.77 & 93.31 & 1.18 & 1.55 \\
        	 MBN(cos)& 94.92 & 95.37 & 94.28 & 94.53 & 94.06 & 95.09 & 94.84 & 93.58 & \textbf{94.58} & \textbf{0.59} & \textbf{1.39} \\ \hline
         Arcface \cite{deng2018arcface} & 96.03 & 95.90 & 93.57 & 93.36 & 93.48 & 95.23 & 94.28 & 93.28 & 94.39 & 1.16 &1.69 \\
         MBN(arc)& 95.93 & 95.57 & 94.28 & 94.77 & 94.57 & 95.57 &  94.21 & 94.46  & \textbf{94.92} & \textbf{0.67} & \textbf{1.42} \\ \hline 	
	\end{tabular}}
    \end{center}
\end{table*}

\textbf{Results on BUPT-Balancedface dataset.} We also compare our MBN with Softmax, Cosface \cite{wang2018cosface} and Arcface \cite{deng2018arcface} on BUPT-Balancedface. We train the models on Balancedface-4 and Balancedface-8, and then evaluate them on IDS-4 and IDS-8 respectively, as shown in Table \ref{balancedface} and Table \ref{balancedface-8bin}. First, with balanced training, Softmax, Cosface and Arcface indeed obtain more balanced performances compared with trained on biased data. So training models on datasets well distributed across all skin tone groups can help to reduce face matcher vulnerabilities on specific cohorts to some extent. For instance, Arcface trained on Balancedface-8 decreases STD from 1.52 to 1.16 compared with trained on Globalface-8. Second, the results obtained by our MBN outperform all compared approaches by improving the fairness metrics in performances across skin tone bins. For example, compared with Arcface \cite{deng2018arcface}, MBN(arc) reduces the SER from 1.65 to 1.37 and reduces the standard deviation by 47\% on IDS-4. Combining our debiasing algorithm and balanced data, we can obtain the fairest performance.

\begin{figure}
\centering
\subfigure[Margins of MBN(arc)]{
\label{margin arc} 
\includegraphics[width=3.8cm]{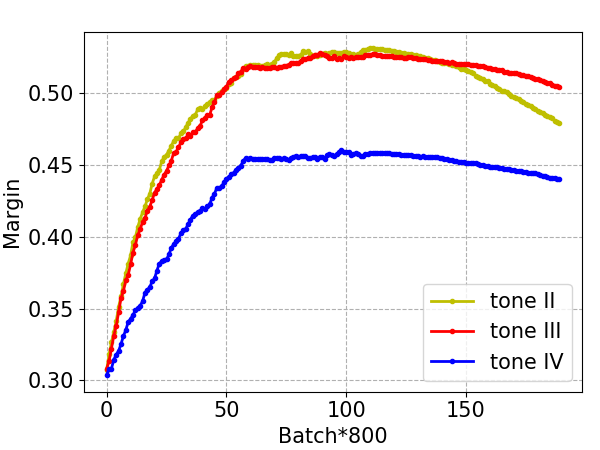}}
\subfigure[Margins of MBN(cos)]{
\label{margin cos} 
\includegraphics[width=3.8cm]{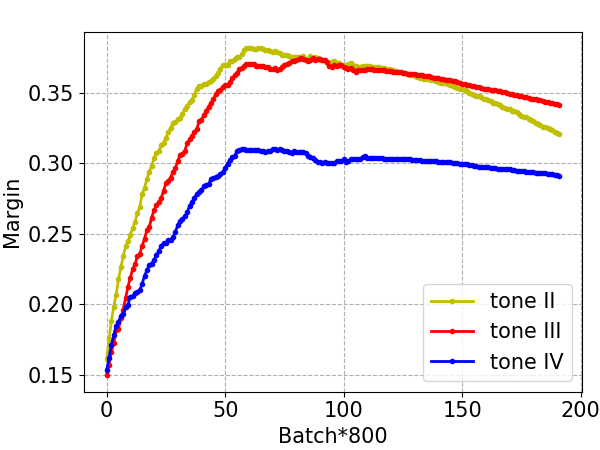}}
\subfigure[Skewness of MBN(arc)]{
\label{bias arc} 
\includegraphics[width=3.8cm]{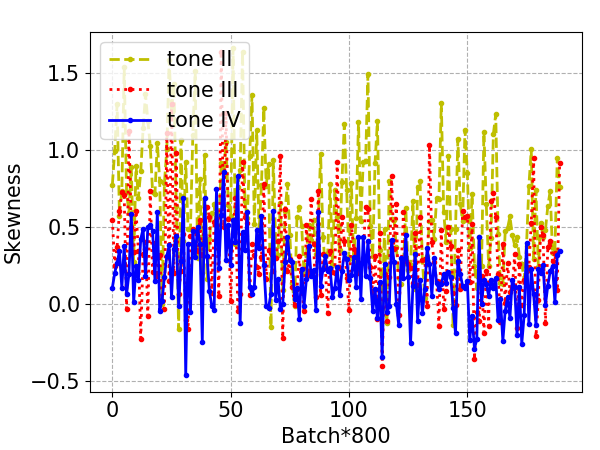}}
\subfigure[Skewness of MBN(cos)]{
\label{bias cos} 
\includegraphics[width=3.8cm]{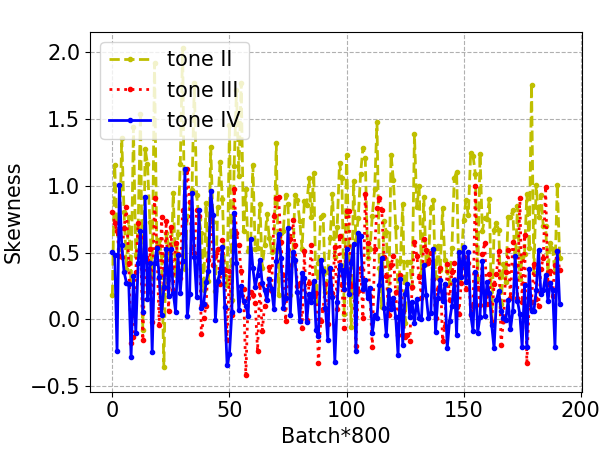}}
\subfigure[Accuracies of MBN(arc)]{
\label{acc arc} 
\includegraphics[width=3.8cm]{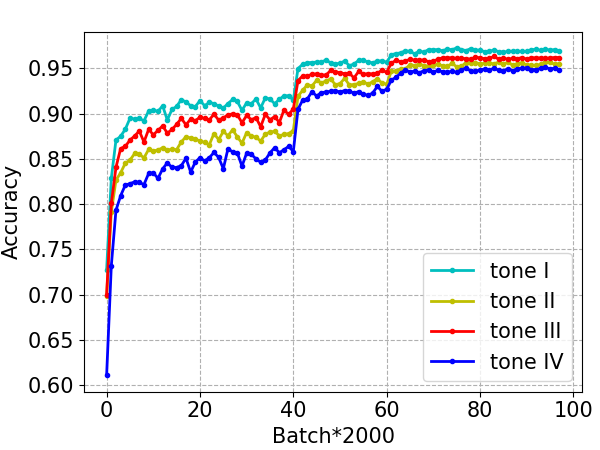}}
\subfigure[Accuracies of MBN(cos)]{
\label{acc cos} 
\includegraphics[width=3.8cm]{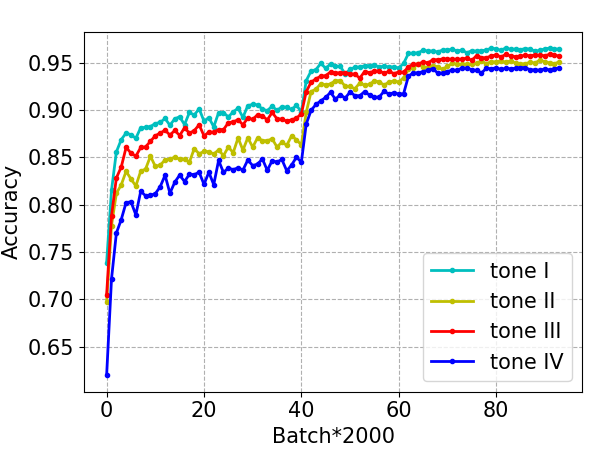}}
\caption{The margins learned for people with skin tone II-IV in (a) MBN(arc) and (b) MBN(cos) trained on Globalface-4. The skewness between lighter-skinned (I) and darker-skinned (II-IV) subjects in (c) MBN(arc) and (d) MBN(cos). The accuracies of different bins in (e) MBN(arc) and (f) MBN(cos).}
\label{margin and bias} 
\end{figure}

\begin{figure}
\centering
\subfigure[Margins of MBN(arc)]{
\label{margin arc} 
\includegraphics[width=3.8cm]{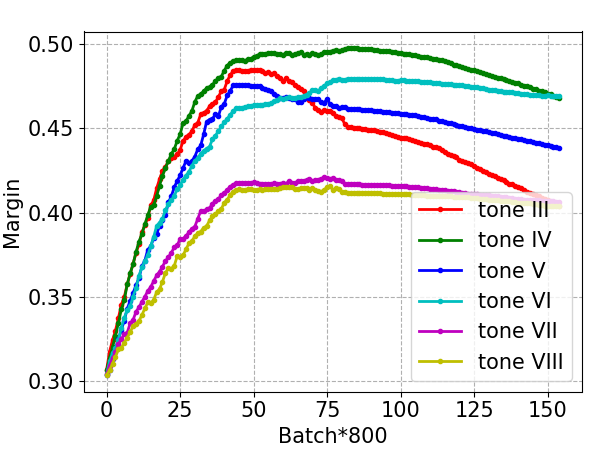}}
\subfigure[Margins of MBN(cos)]{
\label{margin cos} 
\includegraphics[width=3.8cm]{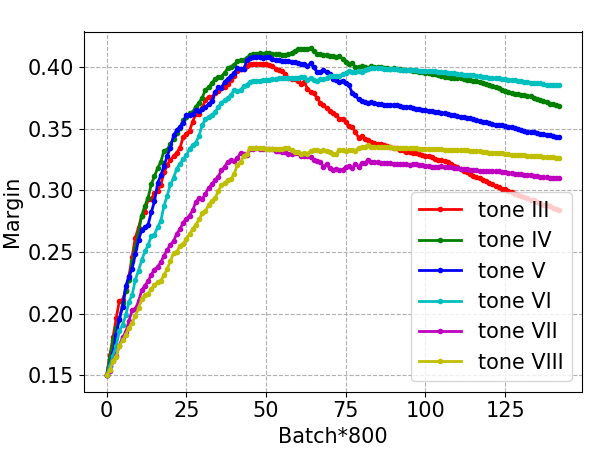}}
\subfigure[Skewness of MBN(arc)]{
\label{bias arc} 
\includegraphics[width=3.8cm]{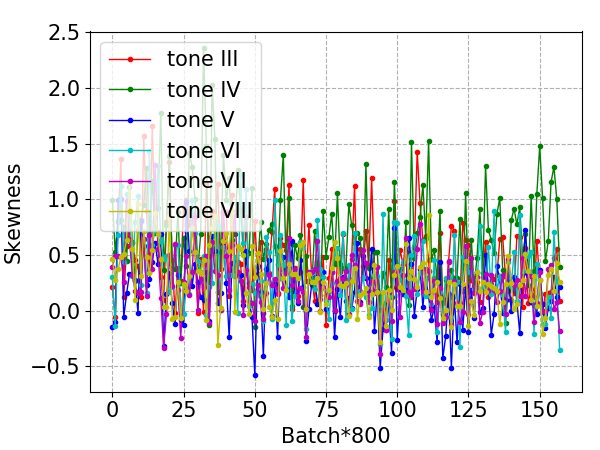}}
\subfigure[Skewness of MBN(cos)]{
\label{bias cos} 
\includegraphics[width=3.8cm]{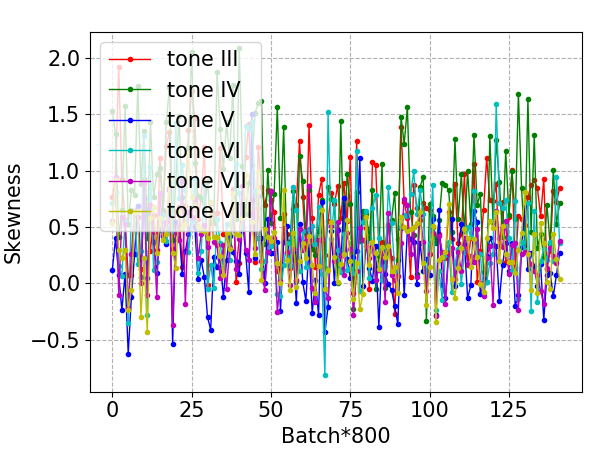}}
\caption{The margins learned for people with skin tone III-VIII in (a) MBN(arc) and (b) MBN(cos) trained on Globalface-8. The skewness between lighter-skinned (I-II) and darker-skinned (III-VIII) subjects in (c) MBN(arc) and (d) MBN(cos). }
\label{margin and bias_8bin} 
\end{figure}

\textbf{Adaptive margin mechanism visualization.} To better understand our MBN, we plot the learned margins, skewness and accuracies of different skin tone bins during the learning process in Fig. \ref{margin and bias} and Fig. \ref{margin and bias_8bin}. First, we can see from Fig. \ref{margin and bias}(a-b) and Fig. \ref{margin and bias_8bin}(a-b) that our method automatically learns larger margins for darker-skinned subjects compared with people with skin tone I. This is consistent with our theoretical analysis that we prefer stricter constraints for dark faces since they are difficult to recognize. Second, the skewness between lighter- and darker-skinned people calculated on meta data by $B^{g}=\frac{1}{n/4}\sum_{i=1}^{n/4}l^{g}_i-\frac{1}{n/4}\sum_{k=1}^{n/4}l^{I}_k$ is shown in Fig. \ref{margin and bias}(c-d) and Fig. \ref{margin and bias_8bin}(c-d), where $g\in\{II,III,IV\}$ and $l^{g}$ and $l^{I}$ can be computed by Eq.  \ref{diff_indian}. On meta data, the skewness is always larger than zero which shows that darker-skinned subjects can not obtain as good intra-class compactness and inter-class discrepancy as Tone-I people do. However, we can see from the results on 4 skin bins shown in Fig. \ref{margin and bias}(c-d), the skewness of Tone-II group is relatively high on meta data even if people with skin tone IV seem more difficult to recognize in test set. Our hypothesis to explain this phenomenon is that there is a little domain shift between meta and test data which is hard to remove entirely. Despite this little shift, our MBN can still learn optimal margin for each bin, leading to more balanced performance on IDS as shown in Fig. \ref{acc arc} and Fig. \ref{acc cos}.

\begin{figure}
\centering
\subfigure[t-SNE on IDS-4]{
\label{fig_v_mbn_a} 
\includegraphics[width=3.8cm]{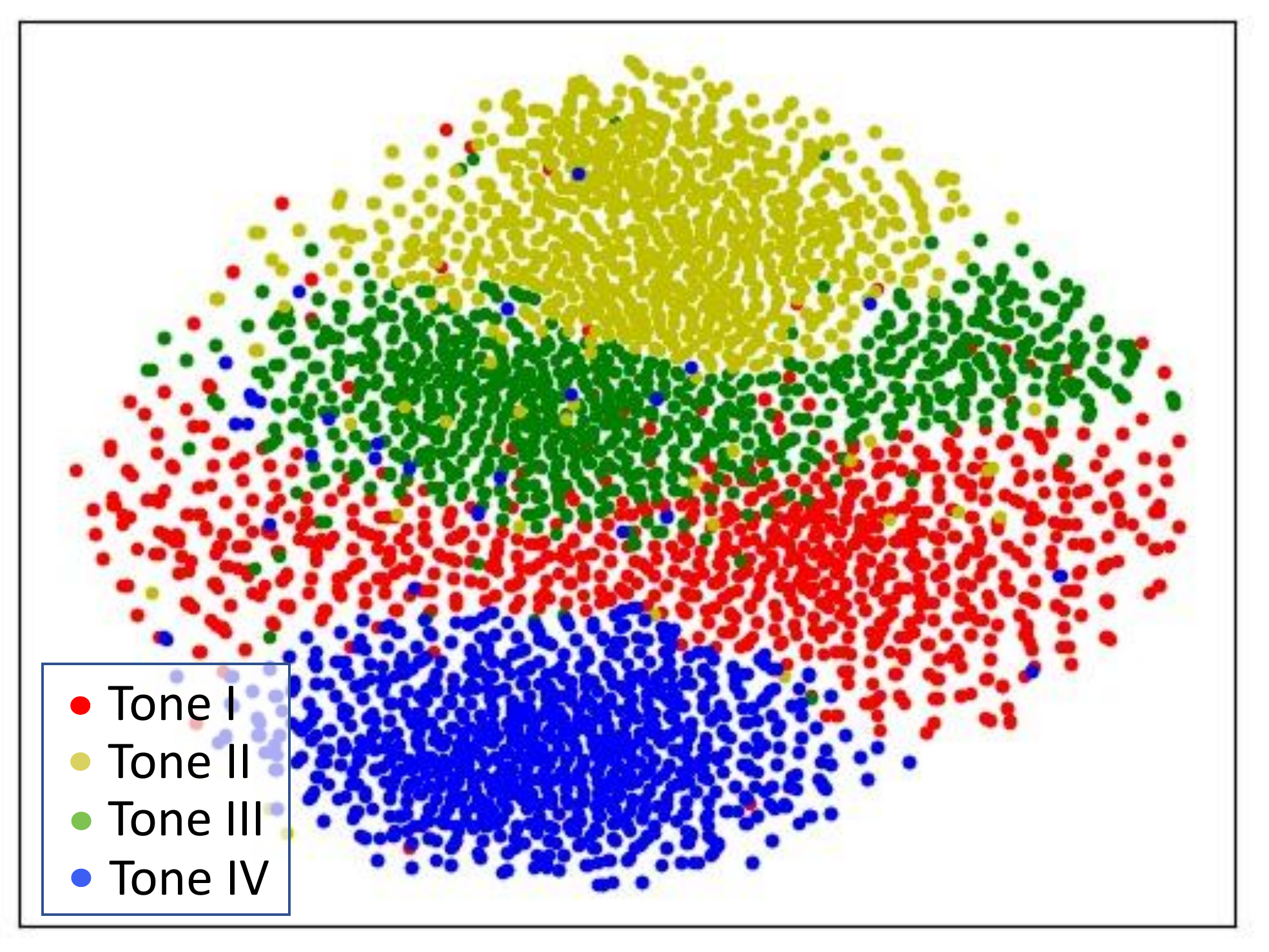}}
\hspace{0cm}
\subfigure[Feature scatter on IDS-4]{
\label{fig_v_mbn_b} 
\includegraphics[width=3.8cm]{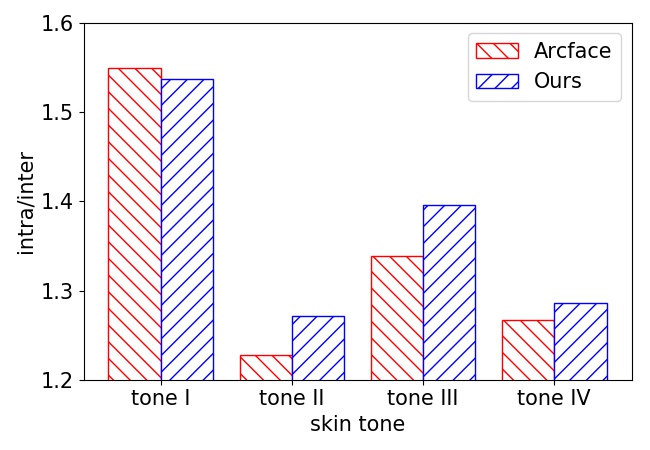}}
\caption{ (a) The feature space and (b) feature scatter of our MBN(arc) model trained on Globalface-4.}
\label{fig_visual_mbn} 
\end{figure}

\textbf{Feature visualization.} Similar to Fig.  \ref{fig4a}, we extract the features by our MBN(arc) and visualize them in Fig. \ref{fig_v_mbn_a}. We can find that there are still domain gaps between different skin tone bins. It is reasonable because our MBN learns fairer representations through balancing the feature scatter across groups instead of domain adaptation or feature disentanglement. The skin tone information is still embedded in the feature space resulting in four separated clusters. Moreover, we also show the feature scatter of Arcface and MBN(arc) in Fig. \ref{fig_v_mbn_b}. The feature scatter is computed by intra-class scatter divided by inter-class scatter, where intra-class scatter refers to the mean of cosine similarities between features and their corresponding feature centres and inter-class scatter refers to the mean of cosine similarities between embedding feature centres. According to the definition, larger feature scatter means better performance. From figure, we can see that feature scatter of Arcface is biased drastically towards subjects with skin tone I while our MBN improves the feature scatter of darker-skinned subjects leading to balanced performance.

\textbf{Tolerance to different variations.}  As we proved in Section \ref{section}, large variations will result in a higher bias across different skin tone groups. Therefore, we measure the influence of illumination, occlusion, Gaussian noise, pose, age and gender on our debiasing method to validate its robustness. As shown in Fig. \ref{variation}, compared with Arcface model, our MBN can consistently improve the performances of darker-skinned subjects and present less bias with respect to skin tone under different conditions, even if the variations are extremely large.
Moreover, our MBN shows more excellent debiasing ability under pose and age variations, especially on larger-pose and younger cohorts which are more difficult to recognize. Compared with Arcface model, MBN reduces the accuracy gap between Tone-I and Tone-IV subjects from 6.58\% to 1.97\% on large-pose group (45$^{\circ}$-90$^{\circ}$) and reduces the gap from 6.83\% to 3.27\% on cohorts with age between 0 to 25. Besides, the results in Fig. \ref{v-gender1} and \ref{v-gender2} show that our debiasing method also performs fairly on different gender cohorts. The skewness with respect to skin tone is reduced from 1.74\% to 0.39\% on males, and from 4.83\% to 3.70\% on females.

\begin{figure}[htbp]
\centering
\includegraphics[width=7.5cm]{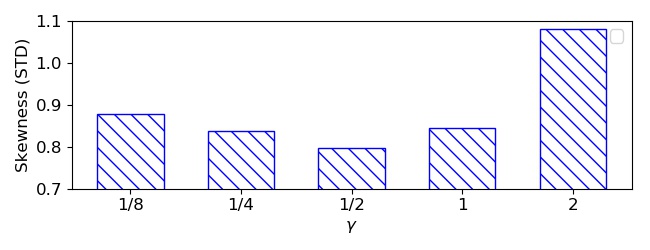}
\caption{ The effect of hyper-parameter $\gamma$ on debiasing ability of our method (STDs of accuracies of different bins are showed).}
\label{min-max coefficient}
\end{figure}

\textbf{Parameter sensitivity.} The hyper-parameter $\gamma$ in Eqn. \ref{diff_indian} is a tradeoff parameter that balances positive-pair distance and negative-pair distance. It affects the measurement of fairness in meta skewness loss and so the debiasing performance. We studied this parameter by setting it to different values and checking the debiasing performance. A ResNet-34 is trained using Globalface-4 dataset and the results tested on IDS-4 are shown in Fig. \ref{min-max coefficient}. We observe that bias first decreases and then increases as $\gamma$ varies from 0.125 to 2 and our MBN performs best when $\gamma$ = 0.5. This suggests that positive-pair distance contributes more in measuring fairness in meta skewness loss. Actually, we find, through experiments, that intra-class compactness in Arcface model is indeed more biased across skin tone bins compared with inter-class discrepancy, so we should focus more on the positive-pair distance and less on the negative-pair distance in debiasing process.

\textbf{Comparison with other adaptive margin methods.} RL-RBN \cite{wang2019mitigate} and M-RBN \cite{wang2019mitigate} published in CVPR'20 are closely related methods with ours, which adopt a similar adaptive margin mechanism. M-RBN utilizes different fixed margins for different skin tone bins inversely proportional to their quantity; and RL-RBN formulates the process of finding the optimal margins for darker-skinned subjects as a Markov decision process and employs deep reinforcement learning to learn margin polices. We show the compared results in Table \ref{worldface} and Table. \ref{balancedface}.  Although M-RBN \cite{wang2019mitigate} can improve the fairness of model compared with Cosface \cite{wang2018cosface} and Arcface \cite{deng2018arcface} benefitting from different margins, the performance of people with skin tone II is always a drag on fairness. This is because bias is a complex problem in which the quantity is not only fact affecting out-of-balance accuracy. Although the number of Tone-II people is much larger than that of people with skin tone III and IV in Globalface-4, Tone-II group still needs a larger margin. Moreover, our MBN obtains more uniform performance than M-RBN and RL-RBN which shows the superiority of our algorithm. RL-RBN utilizes deep Q-learning and makes the search space of margin $\mathcal{M}$ to be discrete by assuming $\mathcal{M}=\{m_1, m_2, m_3, m_4\}$. For example, the optimal margin can only be searched from a subspace $\mathcal{M}_s=\{0.3,0.4,0.5,0.6\}$ in RL-RBN(arc). Our MBN uses gradient based optimization and can conduct a continuous search in the whole space, which enables more suitable margin parameters for darker-skinned people.

\begin{table}[htbp]
    \setlength{\abovecaptionskip}{0cm}
	\setlength{\belowcaptionskip}{-0.2cm}
    \caption{Verification accuracy (\%) of other debiasing methods trained with Arcface loss ([BUPT-Globalface-4, ResNet34, Arcface]).}
    \label{other}
	\begin{center}
	\footnotesize
    \setlength{\tabcolsep}{1.1mm}{
	\begin{tabular}{c|cccc|c|cc}
		\hline
         \multirow{2}{*}{Methods} & \multicolumn{4}{c|}{IDS-4}  & \multirow{2}{*}{Avg} & \multicolumn{2}{c}{Fairness} \\
         & \uppercase\expandafter{\romannumeral1} & \uppercase\expandafter{\romannumeral2} & \uppercase\expandafter{\romannumeral3} & \uppercase\expandafter{\romannumeral4} & & STD & SER \\ \hline \hline
         Arcface \cite{deng2018arcface} & 97.37 & 94.55 & 95.68 &  93.87 & 95.37 & \textcolor{red}{1.53} & \textcolor{red}{2.33} \\
         Re-weight \cite{ren2018learning} & 96.35 & 94.25 & 95.32 &  93.48 & \textcolor{red}{94.85} & 1.25 & 1.79 \\
         Adversarial \cite{alvi2018turning} & 96.63 & 94.17 & 95.27 &  93.70 & 94.94 &  1.30 & 1.87 \\
         MBN(arc)& 96.87 & 95.63 & 96.20 &  95.00 & \textbf{95.93} & \textbf{0.80} & \textbf{1.59} \\ \hline 	
	\end{tabular}}
    \end{center}
\end{table}

\textbf{Comparison with other debiasing methods.} To address the class-imbalance and bias issue, sample re-weighting \cite{ren2018learning,shu2019meta} and adversarial attribute removal methods \cite{alvi2018turning,ganin2014unsupervised} have been exploited and achieved improved performance in some tasks, e.g., object and gender classification. Here, we also compare our MBN with these methods in Table \ref{other}. All comparison methods are trained with Arcface loss \cite{deng2018arcface}  on Globalface-4 dataset.

From the results, we can find that our MBN is superior to all compared methods. The reasons may be as follows. First, there is a tradeoff between model identification performance and model fairness in adversarial attribute removal method. Such identity feature contained little skin-tone information could undermine the recognition competence since skin tone information is a part of identity-related facial appearance. Therefore, in order to maintain satisfactory model identification performance, fairness cannot be improved significantly. Second, sample re-weighting method just makes the network pay more attention to darker-skinned people during training by imposing larger weights on their losses, but actually, this improvement of network generalization on darker-skinned people is very limited since face recognition is a fine-grained and open-set classification problem. However, our MBN can adjust the feature scatter in feature space by learning optimal margins for darker-skinned groups such that intra-class compactness and inter-class discrepancy of darker-skinned subjects can be improved significantly leading to balanced performance.

\textbf{Effectiveness on age and gender bias.} For age bias, we can see from Fig. \ref{v-age1} that Arcface performs best on 25-50 years old cohort, followed by 50-100 age cohort, and worst on younger cohort, which proves that age bias also exists in face recognition algorithms. Compared with Arcface, our MBN can successfully reduce age bias. For example, on Tone-VI bin of IDS-4, Arcface model obtains the accuracies of 87.22\%, 93.58\% and 91.21\% on young, middle-age and old cohorts, respectively, presenting a standard deviation of 3.21; while our MBN decreases the standard deviation from 3.21 to 0.94 and achieves the accuracies of 92.44\%, 94.28\% and 92.96\% on different age cohorts. For gender bias, Arcface always performs worse on females, especially darker-skinned females, as shown in Fig. \ref{v-gender1}, which proves the existence of gender bias. Compared with Arcface, our MBN has a little effect on mitigating gender bias. For example, on Tone-I bin, Arcface model obtains the accuracies of 95.82\% on males and 95.16\% on females, presenting an accuracy gap of 0.66; while our MBN achieves the accuracies of 95.82\% and 95.24\%, decreasing the accuracy gap from 0.66 to 0.58. However, the improvement of fairness with respect to gender is limited. It is reasonable because MBN is designed for bias with respect to skin tone by learning adaptive margins for different skin tone bins without considering gender.

\section{Conclusion}

Considering that the problem of bias with respect to skin tone is yet to be comprehensively studied, we have done the first step for this bias of face recognition and create a benchmark for it. Our IDS database encourages FR algorithms to be fairly evaluated and compared on different skin tone groups. Through experiments on our IDS, we first prove that the deep models trained on the current benchmarks do not perform well on darker-skinned faces and that this bias comes from both data and algorithm aspects. Then, we provide two large-scale training datasets, i.e., BUPT-Globalface and BUPT-Balancedface, to remove bias from data aspect.  
Finally, a meta balanced network is proposed to alleviate bias and learn more balanced features from algorithm aspect. The comprehensive experiments prove the potential and effectiveness of our MBN on reducing bias with respect to skin tone.
However, our MBN should be trained on skin tone aware training datasets and requires access to the sample-level protected attribute, i.e., skin tone label, during training. Hence, one future trend is to investigate more elegant debiasing algorithm such that a balanced model can be learned without the explicit usage of any demographic attributes.

\ifCLASSOPTIONcompsoc
  \section*{Acknowledgments}
\else
  \section*{Acknowledgment}
\fi

This work was supported in part by the National Natural Science Foundation of China under Grants No. 61871052 and BUPT Excellent Ph.D. Students Foundation CX2020207.

\ifCLASSOPTIONcaptionsoff
  \newpage
\fi

{
\bibliographystyle{IEEEtran}
\bibliography{egbib}
}

\begin{IEEEbiography}[{\includegraphics[width=1in,height=1.25in,clip,keepaspectratio]{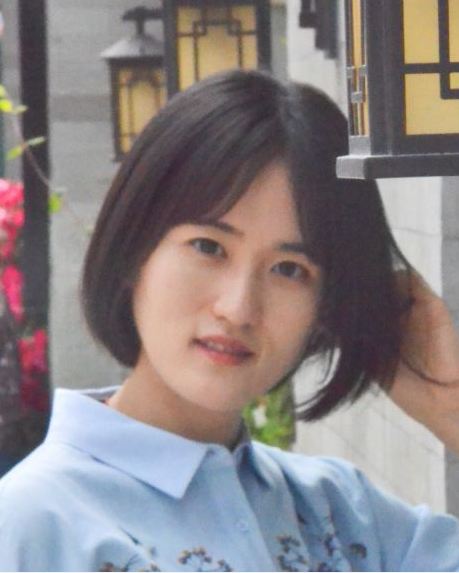}}]{Mei Wang}
received the B.E. degree in information and communication engineering from the Dalian University of Technology (DUT), Dalian, China, in 2013 and received M.E. degree in communication engineering from the Beijing University of Posts and Telecommunications (BUPT), Beijing, China, in 2016. From September 2018, she is a Ph.D. student in school of Artificial Intelligence of BUPT. Her research interests include computer vision, with a particular emphasis in face recognition, transfer learning and AI fairness.
\end{IEEEbiography}

\begin{IEEEbiography}[{\includegraphics[width=1in,height=1.25in,clip,keepaspectratio]{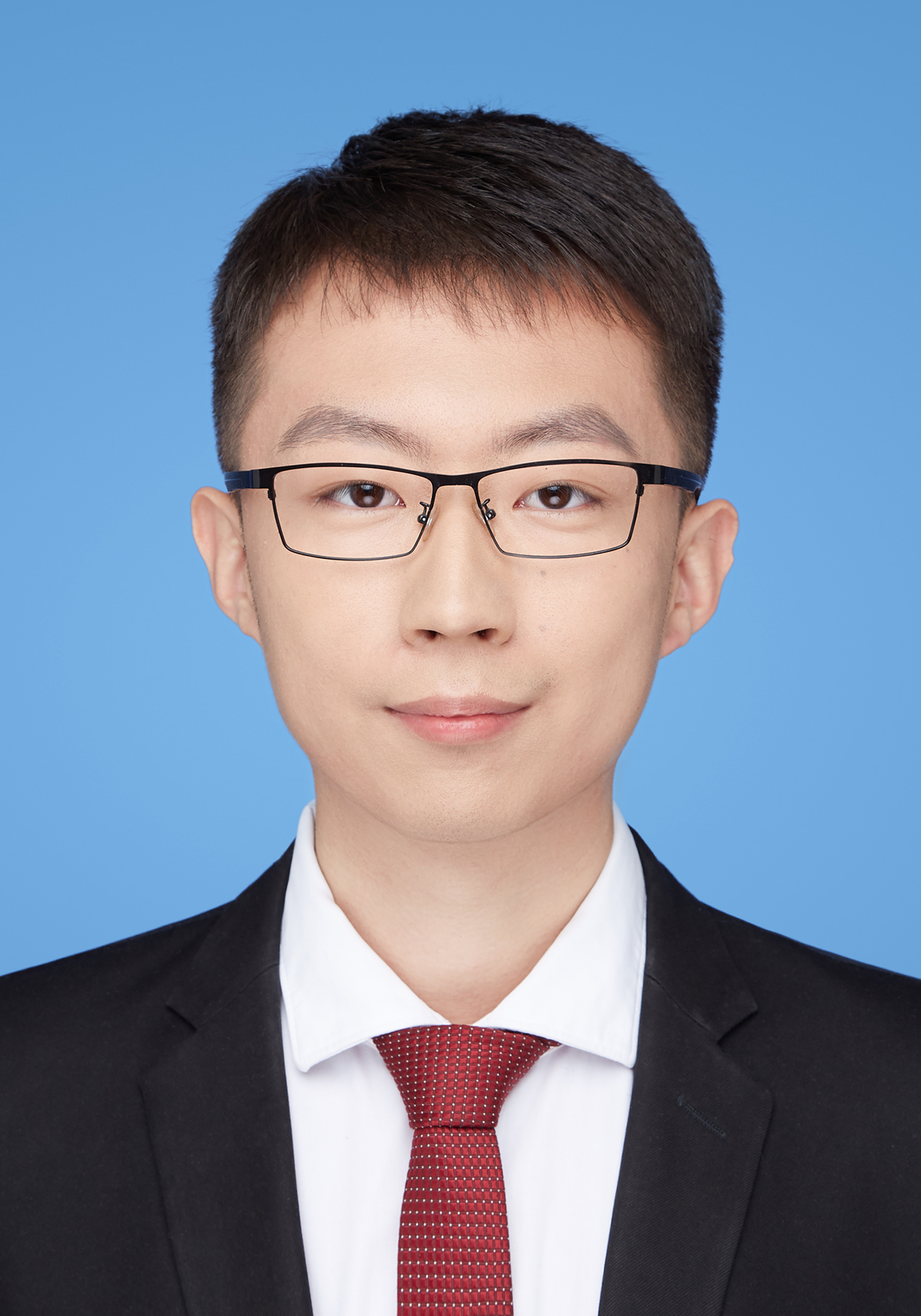}}]{Yaobin Zhang}
was born in Beijing, China, in 1996. He received his B.S. degree in Communication Engineering from Beijing University of Posts and Telecommunications, Beijing, China, in 2019. He is currently pursuing the M.S. degree in Information and Communication Engineering with Beijing University of Posts and Telecommunications. His research interests include deep learning and computer vision.
\end{IEEEbiography}

\begin{IEEEbiography}[{\includegraphics[width=1in,height=1.25in,clip,keepaspectratio]{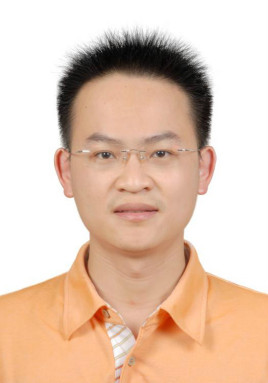}}]{Weihong Deng}
received the B.E. degree in information engineering and the Ph.D. degree in signal and information processing from the Beijing University of Posts and Telecommunications (BUPT), Beijing, China, in 2004 and 2009, respectively. From Oct. 2007 to Dec. 2008, he was a postgraduate exchange student in the School of Information Technologies, University of Sydney, Australia. He is currently a professor in School of Artificial Intelligence, BUPT. His research interests include trustworthy biometrics and affective computing, with a particular emphasis in face recognition and expression analysis. He has published over 100 papers in international journals and conferences, such as IEEE TPAMI, TIP, IJCV, CVPR and ICCV. He serves as area chair for major international conferences such as IJCB, FG, IJCAI, ACMMM, and ICME, guest editor for IEEE Transactions on Biometrics, Behavior, and Identity Science, and Image and Vision Computing Journal, and the reviewer for dozens of international journals and conferences. His Dissertation was awarded the outstanding doctoral dissertation award by Beijing Municipal Commission of Education. He has been supported by the programs for New Century Excellent Talents and Young Changjiang Scholar by Ministry of Education.
\end{IEEEbiography}



\enlargethispage{-5in}

\end{document}